\documentclass[twocolumn]{main}

\paperstyle{fancy}
\papercolor{green}

\DeclareRobustCommand{\method}{{\fontfamily{lmtt}\upshape\selectfont\textbf{STRIDE}}\xspace}
\newtheorem{proposition}{Proposition}[section]
\usepackage{capt-of}
\usepackage{dblfloatfix}
\usepackage{flushend}
\usepackage{needspace}
\definecolor{strideTableHead}{HTML}{F6E3DF}
\definecolor{strideTableOurs}{HTML}{FDF8E3}
\definecolor{strideTableStripe}{gray}{0.95}
\definecolor{strideTableGain}{HTML}{A7441C}
\definecolor{strideTableLoss}{HTML}{006E87}
\definecolor{strideTableTie}{HTML}{595959}
\newcommand{\strideDelta}[2]{\textsuperscript{\textcolor{#1}{\normalfont #2}}}
\makeatletter
\newcommand{\strideband}[2]{\noalign{\begingroup
  \dimen@=\dimexpr\ht\@arstrutbox+\dp\@arstrutbox\relax
  \multiply\dimen@ by #1\relax
  \ifnum#1=2 \advance\dimen@ by \dimexpr\aboverulesep+\cmidrulewidth+\belowrulesep\relax\fi
  \color{#2}\hrule height\dimen@\vskip-\dimen@
  \endgroup}}
\makeatother
\newcommand{\tabsetup}{\footnotesize\setlength{\tabcolsep}{3.2pt}\renewcommand{\arraystretch}{1.15}\setlength{\aboverulesep}{0pt}\setlength{\belowrulesep}{0pt}\arrayrulecolor{black}}
\newcommand{\pre}{\textsc{pre}}
\newcommand{\edit}{\textsc{edit}}

\newcommand{\maketitlesupplementary}{%
  \clearpage
  \twocolumn[%
    \noindent{\Large\sffamily\bfseries\color{accentcolor}Appendix\par}%
    \vspace{0.25em}%
    {\color{rulecolor!70}\rule{\textwidth}{0.65pt}}\par%
    \vspace{0.6em}%
  ]%
  \addcontentsline{toc}{section}{Appendix}%
}

\title{Say It Differently, Move the Same: Trustworthy Motion Generation for Language-Driven Humanoids}

\author[1,2]{Wenshuo Chen}
\author[1,\ast]{Zhuo Chen}
\author[2]{Honglin Chu}
\author[1]{Shaofeng Liang}
\author[3]{Lei Wang}
\author[4]{Shang Gao}
\author[2]{Jianfei Song}
\author[2]{Shiming Li}
\author[5]{Xiucheng Wang}
\author[6]{Le Yang}
\author[1]{Bowen Tian}
\author[1]{Jiemin Wu}
\author[1]{Songning Lai}
\author[1,\dagger]{Yutao Yue}

\affiliation[1]{The Hong Kong University of Science and Technology (Guangzhou)}
\affiliation[2]{LimX Dynamics}
\affiliation[3]{Griffith University}
\affiliation[4]{University of Macau}
\affiliation[5]{Xi'an University of Electronic Science and Technology}
\affiliation[6]{Sun Yat-sen University}

\contribution[\ast]{Core contributor}
\contribution[\dagger]{Corresponding author}
\correspondence{\email{yutaoyue@hkust-gz.edu.cn}}

\abstract{
Trustworthy humanoids must follow a command however it is phrased. Changes in wording or syntax can leave a physical robot stalled in place or cause a fall during simulated tracking. How can a generator preserve an action when its wording changes and alter that action when its meaning changes? \method{} combines motion-conditioned attribution with relation-conditioned trajectory anchoring. During training, it pools token features by reconstruction sensitivity (Grad$\times$Input) on a shared noisy motion state to align paraphrases of different lengths and word orders. At a state reached under the paired command, the student learns to match its own original-command prediction for a paraphrase (PRE) or a frozen reference model's edited-command prediction for a semantic edit (EDIT). On ECHO Lite, \method{} raises motion embedding similarity across paraphrases (RepCos) from 0.571 to 0.835. Compared with same-input on-policy distillation under matched training, \method{} reduces the motion distribution distance across rewordings (FID$_D$) by 30.0\% and raises paired MuJoCo tracking completion from 91.7\% to 92.9\%. Adding edit anchoring raises paired left/right accuracy from 54.9\% to 76.9\%. On a physical Unitree G1, \method{} crosses its forearms and walks forward under reworded commands. ECHO Lite stalls on the walking task. Inference retains the original sampler and requires no relation labels.
}

\hypersetup{pdfauthor={Wenshuo Chen, Zhuo Chen, Honglin Chu, Shaofeng Liang, Lei Wang, Shang Gao, Jianfei Song, Shiming Li, Xiucheng Wang, Le Yang, Bowen Tian, Jiemin Wu, Songning Lai, Yutao Yue}}

\begin{document}
\maketitle
\begin{figure*}[t]
  \centering
  \includegraphics[width=\textwidth]{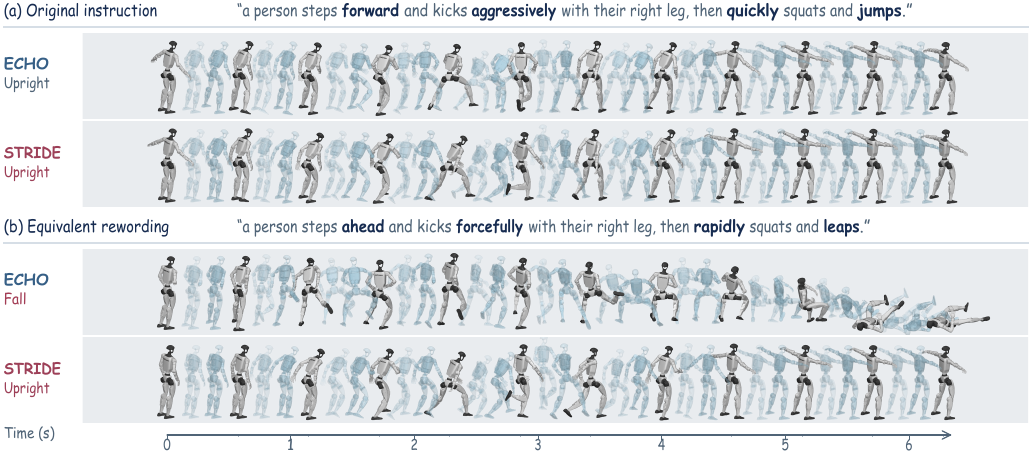}
  \caption{\textbf{G1 tracking in simulation.} After rewording, ECHO falls and \method{} stays upright. Lighter poses mark intermediate times.}
  \label{fig:teaser}
\end{figure*}
\section{Introduction}

Text-to-motion models can synthesize a range of human movements from free-form descriptions~\cite{tevet2023human,Guo_2024_CVPR}.
Recent work brings this capability to humanoid robots by coupling a language-conditioned motion generator with a whole-body tracking controller~\cite{echo,textop,fromw1}.
The generator specifies a reference trajectory, and the controller attempts to realize it under the dynamics of the robot.
A motion can thus be judged by its naturalness, its agreement with the command, and its feasibility under tracking.
Even a motion that meets all three criteria leaves a question open: would the robot carry out the same intent if the user expressed it differently?

Consider the two commands in Figure~\ref{fig:teaser}.
The first asks the robot to step forward, kick with its right leg, squat, and jump.
In simulation, a Unitree G1 tracking the motion generated by ECHO completes the sequence upright.
The second command substitutes expressions such as \emph{step ahead} for \emph{step forward} and \emph{leap} for \emph{jump}.
The sequence, the active leg, and the manner of movement are preserved.
With the same controller, the robot falls while tracking the reworded reference.
A test of the original command would record a success. The paired executions expose its dependence on wording.

What should remain consistent across the two commands?
A language instruction rarely specifies a unique trajectory. A person asked to walk quickly may take longer strides, increase their cadence, or do both.
We call humanoid motion generation \emph{faithful} when the constraints specified by the user are preserved across equivalent wordings and carried through execution, while the generator retains the freedom to vary how those constraints are satisfied.
This distinction matters for evaluation. A change in stride length across random samples can reflect legitimate diversity. A systematic shift caused by replacing \emph{quickly} with \emph{rapidly} would indicate sensitivity to wording.
We therefore compare wordings under shared sampling noise and separately measure variation across samples of a fixed command.

\begin{figure}[!t]
  \centering
  \includegraphics[width=.99\columnwidth]{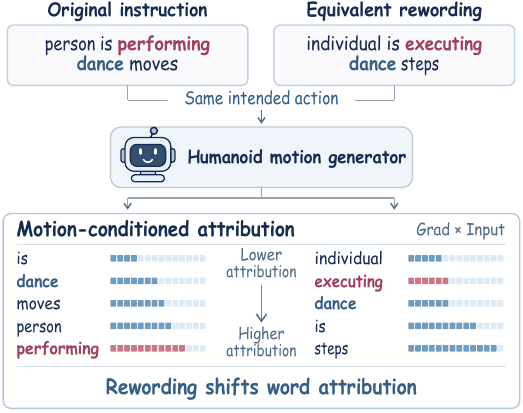}
  \caption{\textbf{Schematic of word-level attribution.} Bar lengths illustrate how a dance description can shift weight between words. Each caption has its own word positions.}
  \label{fig:intro-attribution}
\end{figure}

How does the generator itself read an instruction?
We introduce a \emph{motion-conditioned attribution diagnostic} that characterizes this reading through the generator's response to its text input.
We freeze the generator and present two instructions with the same target motion at the same noisy state and timestep.
For each token, we multiply its input embedding elementwise by the gradient of the motion reconstruction score and sum the absolute values across embedding dimensions.
We derive pooling weights from these scores to obtain a motion-conditioned representation for each instruction, even when the two instructions differ in length or vocabulary.
Figure~\ref{fig:intro-attribution} illustrates how words can carry different reconstruction evidence across two descriptions of the same dance.
In a gradient-norm probe of ECHO Lite, mean pooled distance was greater for lexical rewordings than for direction edits. The two groups contained different motions.

\begin{figure*}[t]\centering\includegraphics[width=\textwidth]{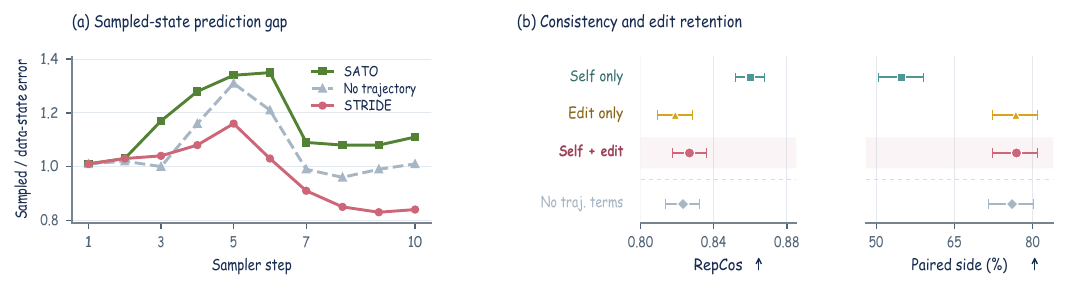}
\caption{\textbf{Wording sensitivity along the sampler.} (a) A ratio above one means a larger prediction gap on sampled states. (b) Edit anchoring improves paired-side accuracy over self-only. Bars show 95\% family intervals.}\label{fig:focus}\end{figure*}

This observation suggests aligning motion-relevant evidence across equivalent instructions.
Agreement alone admits a simple solution: a generator that pays little attention to language can agree across many wordings.
An edit from \emph{walk slowly} to \emph{walk quickly} identifies a distinction the generator must retain. Both outputs should remain walks, and the second should be faster.
Edits thus provide a behavioral constraint on what alignment may discard.
There is also a question of where to enforce that constraint. During diffusion sampling, each denoising prediction helps determine the state on which the next prediction is made.
We find that rewording can induce larger prediction discrepancies on these sampled states than on noised training motions at the same timestep.
This motivates supervising motion predictions along the generator's own trajectory, with a target chosen according to what changed in the instruction.

We introduce \textbf{\method{}}, a fine-tuning method that implements this principle.
During training, we run the current sampler under the paired instruction and stop at an intermediate state.
At that state, an equivalent rewording is trained to match the current generator's prediction for the original command. An edit receives the prediction of a frozen reference generator that sees the edited command.
Attribution-weighted alignment provides additional supervision for equivalent wordings, and learned registers replace sentence-dependent padding features.
Inference uses the original sampler and requires a single instruction.
On ECHO Lite, \method{} improves motion consistency across model-generated, human-written, and rule-based rewordings while retaining measured responses to edits in direction, speed, and action.
The gain persists through G1 tracking in MuJoCo. On model-generated rewordings, executed-motion similarity rises from 0.792 with SATO~\cite{sato} to 0.836 with \method{}.

Our contributions are threefold.
\begin{itemize}
  \item \textbf{Faithfulness in humanoid motion generation.}
  We formulate faithfulness in terms of the motion constraints an instruction specifies and the variation it permits. Paired generation and tracking tests separate sensitivity to wording from within-command sampling variation and assess whether the requested behavior survives execution.
  \item \textbf{A motion-conditioned attribution diagnostic.}
  We use the generator's local response to text at a common motion state to compare how it reads different instructions. We measure how token weighting separates equivalent from unrelated commands and test its effect by attenuating the highest-ranked words.
  \item \textbf{Relation-conditioned trajectory anchoring.}
  We train on states visited by the generator and select a self-target or a frozen-reference target according to the relation between commands. Matched controls isolate the benefit of this choice for wording consistency, and simulated G1 tracking shows that the gain carries over to executed motion.
\end{itemize}

\section{Related Work}
\label{sec:related}
\begin{figure*}[t]\centering
\includegraphics[width=\linewidth]{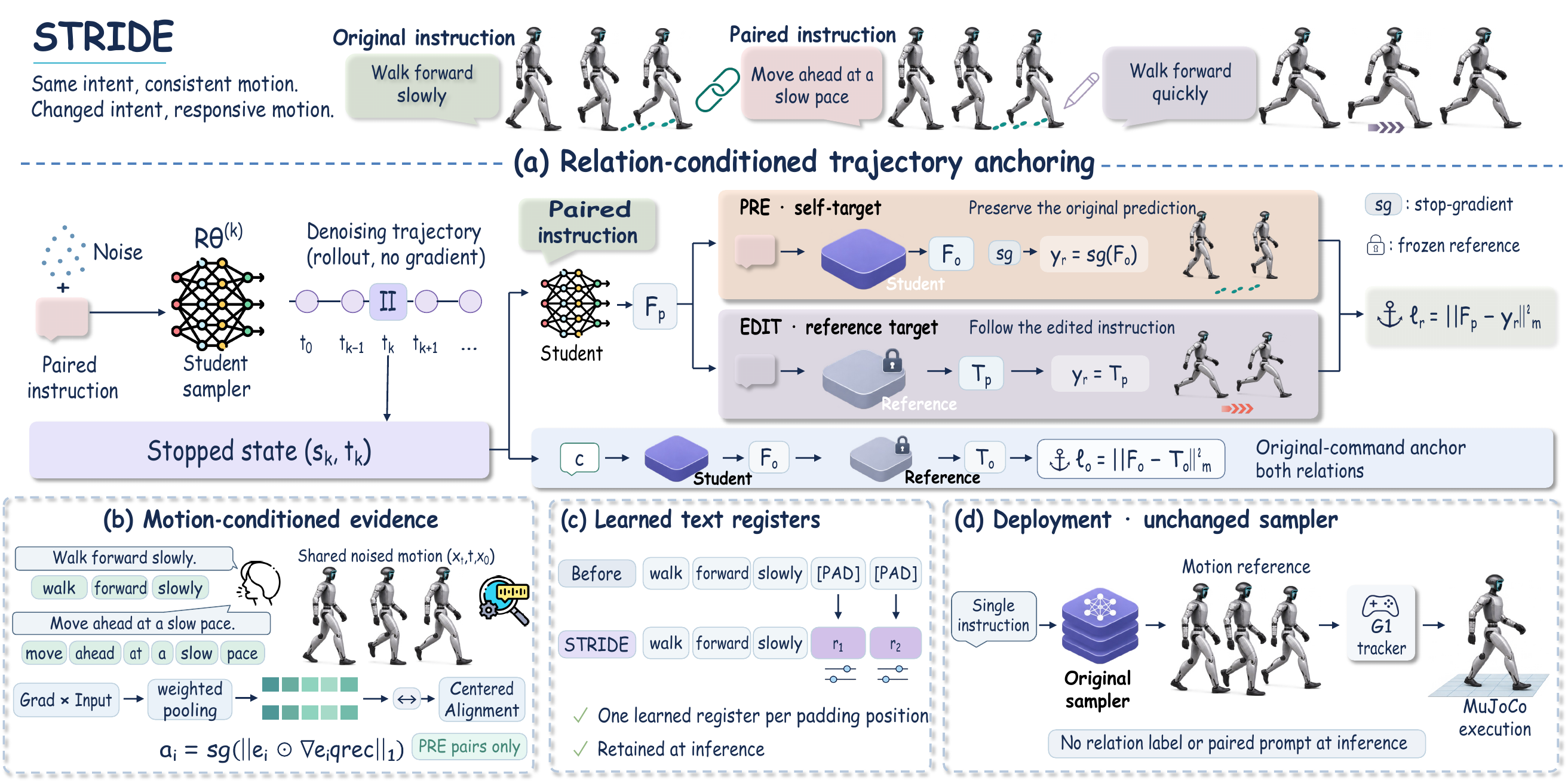}
\caption{\textbf{\method{} training and deployment.} At the same sampler state, a paraphrase follows the student prediction for the original command. An edit follows the reference prediction for the edited command.}
\label{fig:relation}\end{figure*}

\textbf{Language-conditioned motion.} Text-to-motion models use motion--caption pairs~\cite{humanml3d, metrics} to learn continuous~\cite{tevet2023human,motiondiffuse,mld} or discrete~\cite{t2mgpt,Guo_2024_CVPR} motion representations. Physics guidance~\cite{Yuan_2023_ICCV} and language-conditioned whole-body control~\cite{pmlr-v270-jiang25b,ShaoY-RSS-25,jiang2025uniactunifiedmotiongeneration} connect generation to execution.

\textbf{Robustness and attribution.} Prior work adapts CLIP features~\cite{clip,leaf,moclip}, stabilizes attention and predictions~\cite{sato}, guides denoising~\cite{multiact,ahn2025headhunter,ant,tcr}, or selects noise~\cite{winro}. Free-T2M models motion frequencies~\cite{freet2m}. T2MBench evaluates instruction following across diverse prompts~\cite{t2mbench}. We pair commands under shared noise to isolate wording effects on a fixed backbone. Attention weights and input attributions measure different aspects of token influence~\cite{attentionnot,attentionnotnot,integratedgradients,deeplift}. \method{} aligns pooled text representations using motion reconstruction sensitivity.

\textbf{Learning on sampled states.} Diffusion samplers~\cite{ddim,dpmsolver} and distillation~\cite{progressive,consistency} can reduce inference cost. TACD distills a few-step motion generator~\cite{tacd}. On-policy methods supervise learner-generated states~\cite{dagger,gkd,flowopd,cfgopd,danceopd}, whose errors can differ from those on noised data~\cite{exposurebias,exposurebias2}. \method{} keeps the sampler fixed and selects the target by the semantic relation between commands.

\section{The Proposed \method{}}
\label{sec:method}
\subsection{Evidence for the design}
\label{sec:diag}
Equivalent commands can assign reconstruction evidence to different words. Even when a generator detects a change in wording, the trajectory target determines whether the change survives sampling. We examine both stages on ECHO Lite under shared noise.

\textbf{Where wording changes the rollout.} For SATO, equivalent commands produce a larger prediction gap at states visited by the sampler than at matched noised data states (Figure~\ref{fig:focus}(a)). The gap also grows in the base model. Relation-conditioned supervision narrows it on the states used at deployment.

\textbf{Which responses should survive?} Self-anchoring gives the highest wording consistency but also the lowest paired-side accuracy under a left/right edit. Adding an edit target recovers much of that response (Figure~\ref{fig:focus}(b)). The self-only, edit-only and joint models share a trainer and base objective.

\textbf{Which words matter to reconstruction?} Nonlexical positions can absorb attention in language and image generators~\cite{xiao2024attentionsinks,toker-etal-2025-padding}, and vision transformers can use dedicated registers~\cite{darcet2024registers}. In ECHO Lite, padding receives most of the cross-attention mass but little of the reconstruction gradient. We therefore use reconstruction sensitivity to weight the words in each caption and replace sentence-dependent padding features with learned registers.

\textbf{Training overview.} \method{} selects its anchoring target from the command pair's semantic relation. A training tuple $(x_0,M,c,c',r)$ contains a clean motion, its valid-frame mask, an original command, a paired command and a relation $r\in\{\pre,\edit\}$, denoting meaning preservation and semantic editing. At each update we stop a partial rollout, evaluate both commands at that state and select the paired target by $r$.

\subsection{Preliminaries: diffusion and flow}
Let $x_0\in\mathbb R^{L\times D}$ have $L$ padded frames and $D$ motion coordinates. For a residual $U$ of this shape, binary mask $M_\ell$ selects valid frame $\ell$. We use
\begin{equation}
\|U\|_M^2=\frac{1}{Z}\sum_{\ell=1}^{L}\sum_{d=1}^{D}M_\ell U_{\ell d}^2,
\qquad Z=D\sum_{\ell=1}^{L}M_\ell>0.
\label{eq:mask}
\end{equation}
Here $d$ indexes coordinates and $Z$ counts valid scalars. Diffusion time $t\in[0,1]$ indexes a noise path, independently of the physical frame index $\ell$:
\begin{equation}
x_t=\alpha_t x_0+\sigma_t\epsilon,\qquad
\epsilon\sim\mathcal N(0,I).
\label{eq:path}
\end{equation}
$I$ is the $LD$-dimensional identity; padding is masked. The scalar schedules $\alpha_t,\sigma_t$ interpolate from data $(\alpha_0,\sigma_0)=(1,0)$ toward noise. Diffusion~\cite{ddpm} trains a conditional denoiser and samples this path in reverse. We express predictions in clean-motion coordinates: a noise predictor $\epsilon_\theta$ gives $f_\theta(x,t,c)=(x-\sigma_t\epsilon_\theta(x,t,c))/\alpha_t$ where $\alpha_t\ne0$; direct clean-motion prediction uses $f_\theta$ itself. Here $x$ is an input motion state at time $t$, and $\theta$ collects trainable parameters.

Flow matching~\cite{flowmatching} instead fits the path velocity $\dot\alpha_t x_0+\dot\sigma_t\epsilon$, with dots denoting time derivatives. For the linear path $\alpha_t=1-t$, $\sigma_t=t$, a velocity predictor $v_\theta$ gives
\begin{equation}
f_\theta(x,t,c)=x-t\,v_\theta(x,t,c).
\label{eq:flow}
\end{equation}
Both parameterisations therefore predict a clean motion to which the same anchors can be applied. Our experiments instantiate this interface with diffusion. The frozen reference $f_{\mathrm{ref}}$ is a copy of the pretrained generator.

\subsection{A relation selects its target on the trajectory}
Let $\mathcal R_\theta$ be the deployed $K$-step sampler and $\mathcal R_\theta^{(k)}(c',z)$ its partial rollout through step $k$, with $z$ collecting its initial and per-step random noise. Its decreasing time grid is $t_0>\cdots>t_K$. We draw $k$ uniformly from $\{1,\ldots,K-1\}$ and set
\begin{equation*}
s_k=\mathrm{sg}\!\left[\mathcal R_\theta^{(k)}(c',z)\right].
\end{equation*}
The stop-gradient operator $\mathrm{sg}$ preserves values and has zero derivative. The sampler retains any required multistep history; $s_k$ denotes its motion state. At the same $(s_k,t_k)$, abbreviate the conditional predictions as $F_o=f_\theta(s_k,t_k,c)$, $F_p=f_\theta(s_k,t_k,c')$, $T_o=f_{\mathrm{ref}}(s_k,t_k,c)$ and $T_p=f_{\mathrm{ref}}(s_k,t_k,c')$. Subscripts $o,p$ mean original and paired.
\begin{equation}
y_r=\begin{cases}\mathrm{sg}(F_o),&r=\pre,\\ T_p,&r=\edit,\end{cases}
\qquad \ell_r=\|F_p-y_r\|_M^2.
\label{eq:traj}
\end{equation}
An original-command anchor, $\ell_o=\|F_o-T_o\|_M^2$, constrains the moving self-target. For a minibatch, let $\langle\cdot\rangle_r$ average examples of relation $r$ and $\langle\cdot\rangle$ average all examples; an empty relation contributes zero. The objective is
\begin{equation}
\begin{split}
\mathcal L={}&\mathcal L_{\mathrm{base}}
+\lambda_p\langle\ell_r\rangle_{\pre}
+\lambda_e\langle\ell_r\rangle_{\edit}
+\lambda_o\langle\ell_o\rangle .
\end{split}
\label{eq:objective}
\end{equation}
Figure~\ref{fig:relation} shows how both commands use the same sampler state for anchoring. The weights $\lambda_p,\lambda_e,\lambda_o$ control preserve, edit and original anchoring. The shared base objective combines denoising, reference fidelity and unconditional supervision with the evidence alignment below. All student branches share $\theta$.

\subsection{Motion-conditioned attribution and interface grounding}
\label{sec:attr_method}
For each rewording pair, both captions reconstruct the same $x_0$ from the same $(x_t,t)$. With input token embedding $e_i$, define
\begin{align*}
q_{\rm rec}(c)&=-\|f_\theta(x_t,t,c)-x_0\|_M^2,\\
a_i&=\mathrm{sg}\!\left(\|e_i\odot\nabla_{e_i}q_{\rm rec}(c)\|_1\right).
\end{align*}
Let $I_c$ contain all tokens between BOS and EOS, including punctuation, and $\bar a$ be their mean score. For temperature $\tau_a$ and a positive numerical floor $\varepsilon_a$, pooling uses
\begin{align*}
\alpha_i&=\operatorname{softmax}_{i\in I_c}\!\left(\frac{a_i}{\tau_a\max(\bar a,\varepsilon_a)}\right),\\
u(c)&=\sum_{i\in I_c}\alpha_i P(H_i).
\end{align*}
Here $H_i$ is the generator's contextual token feature and $P$ its learned projection. We use stop-gradient prediction~\cite{simsiam} with feature centering~\cite{dino} and maximize cosine similarity between the prediction from one centered pool and the other caption's detached, centered pool, averaging the two directions. The alignment loss updates the text features and pooling head through $u(c)$. Detaching $a_i$ blocks its gradient path through the denoiser. Learned registers follow EOS, outside the pool.

\subsection{Theoretical analysis of trajectory anchors}
\begin{proposition}\label{prop:edit}
At a fixed state, let $\lambda_e,\lambda_o>0$, $\delta_T=\|T_p-T_o\|_M$, and $E=\lambda_e\|F_p-T_p\|_M^2+\lambda_o\|F_o-T_o\|_M^2$. Then
\begin{equation}
\big|\,\|F_p-F_o\|_M-\delta_T\,\big|
\le \sqrt{(\lambda_e^{-1}+\lambda_o^{-1})E}.
\label{eq:response}
\end{equation}
\end{proposition}
The reverse triangle inequality bounds the left side by the sum of the two anchor errors. Weighted Cauchy--Schwarz gives the right side. A reference that barely responds to an edit offers little separation to retain.

\begin{proposition}\label{prop:stability}
Fix a preserve state and a compatible solution $\theta_*$ with $F_p=F_o=T_o$. Let $J_p,J_o$ be the parameter Jacobians of output vectors restricted to valid coordinates and divided by $\sqrt Z$ at $\theta_*$. For positive local coefficients $w_p,w_o$, the semi-gradient of $w_p\ell_r+w_o\ell_o$ has linearisation
\begin{equation}
A=2w_pJ_p^\top(J_p-J_o)+2w_oJ_o^\top J_o.
\label{eq:linear}
\end{equation}
If $w_o>w_p/4$, the symmetric part of $A$ is positive definite on $V=(\ker J_p\cap\ker J_o)^\perp$, the parameter directions affecting these outputs. Consequently, the fixed linearised ordinary gradient update contracts on $V$ for a sufficiently small positive step.
\end{proposition}
Here $\ker$ denotes a nullspace and $\perp$ its orthogonal complement. The coefficients include the minibatch reductions in Eq.~\ref{eq:objective}. The threshold follows by writing $a=J_ph$, $b=J_oh$ for a displacement $h$. Then $h^\top Ah=2w_p\|a-b/2\|_2^2+2(w_o-w_p/4)\|b\|_2^2$, which is positive for nonzero $h\in V$.

\section{Experimental Setup}
\label{sec:setup}
\begin{table*}[t]\centering
\caption{\textbf{Shared-backbone benchmark and G1 tracking.} VLM check (full matches on 200 synonym rewrites, \%) and G1 tracking use motions generated with seed 42. Arrows show changes from ECHO. Orange marks gains, blue losses. Best in bold, second-best underlined.}
\label{tab:main}\label{tab:embodied}\tabsetup
\begin{tabular*}{\textwidth}{@{\extracolsep{\fill}}lllllc@{\hspace{8pt}}lll@{}}
\toprule
\strideband{2}{strideTableHead}
\multirow{2}{*}{\textbf{Method}}  &  \multicolumn{4}{c}{\textbf{Motion generation (mean over 3 seeds)}} & \multirow{2}{*}{\textbf{VLM check $\uparrow$}}  &  \multicolumn{3}{c}{\textbf{G1 tracking (MuJoCo)}} \\
\cmidrule(lr){2-5}\cmidrule(lr){7-9}
\bfseries  &  \textbf{FID $\downarrow$}  &  \textbf{FID$_D$ $\downarrow$}  &  \textbf{RepCos $\uparrow$}  &  \textbf{Edit Flip $\uparrow$}  &  &  \textbf{PIF (\%) $\downarrow$}  &  \textbf{xRepCos $\uparrow$}  &  \textbf{xFlip $\uparrow$} \\
\midrule
\strideband{1}{strideTableStripe}
ECHO~\cite{echo} & 0.0374 & 0.0275 & 0.571 & 0.902 & 63.0 & 3.8 & 0.611 & \underline{0.900} \\\midrule
MultiAct~\cite{multiact} & 0.0389\strideDelta{strideTableLoss}{$\uparrow$.0015} & 0.0274\strideDelta{strideTableGain}{$\downarrow$.0001} & 0.569\strideDelta{strideTableLoss}{$\downarrow$.002} & 0.902\strideDelta{strideTableTie}{0} & 63.5\rlap{\strideDelta{strideTableGain}{$\uparrow$.5}} & 3.8\strideDelta{strideTableTie}{0} & 0.606\strideDelta{strideTableLoss}{$\downarrow$.005} & 0.887\strideDelta{strideTableLoss}{$\downarrow$.013} \\
WINRO~\cite{winro} & 0.0383\strideDelta{strideTableLoss}{$\uparrow$.0009} & 0.0278\strideDelta{strideTableLoss}{$\uparrow$.0003} & 0.511\strideDelta{strideTableLoss}{$\downarrow$.060} & 0.897\strideDelta{strideTableLoss}{$\downarrow$.005} & 64.5\rlap{\strideDelta{strideTableGain}{$\uparrow$1.5}} & 3.6\strideDelta{strideTableGain}{$\downarrow$.2} & 0.559\strideDelta{strideTableLoss}{$\downarrow$.052} & 0.873\strideDelta{strideTableLoss}{$\downarrow$.027} \\
LwF~\cite{lwf} & \textbf{0.0308}\strideDelta{strideTableGain}{$\downarrow$.0066} & 0.0266\strideDelta{strideTableGain}{$\downarrow$.0009} & 0.572\strideDelta{strideTableGain}{$\uparrow$.001} & 0.892\strideDelta{strideTableLoss}{$\downarrow$.010} & 64.0\rlap{\strideDelta{strideTableGain}{$\uparrow$1.0}} & 2.8\strideDelta{strideTableGain}{$\downarrow$1.0} & 0.615\strideDelta{strideTableGain}{$\uparrow$.004} & 0.877\strideDelta{strideTableLoss}{$\downarrow$.023} \\
LEAF~\cite{leaf} & 0.0372\strideDelta{strideTableGain}{$\downarrow$.0002} & 0.0267\strideDelta{strideTableGain}{$\downarrow$.0008} & 0.579\strideDelta{strideTableGain}{$\uparrow$.008} & \underline{0.906}\strideDelta{strideTableGain}{$\uparrow$.004} & 65.5\rlap{\strideDelta{strideTableGain}{$\uparrow$2.5}} & 3.6\strideDelta{strideTableGain}{$\downarrow$.2} & 0.612\strideDelta{strideTableGain}{$\uparrow$.001} & 0.877\strideDelta{strideTableLoss}{$\downarrow$.023} \\
MoCLIP~\cite{moclip} & 0.0388\strideDelta{strideTableLoss}{$\uparrow$.0014} & 0.0280\strideDelta{strideTableLoss}{$\uparrow$.0005} & 0.576\strideDelta{strideTableGain}{$\uparrow$.005} & \underline{0.906}\strideDelta{strideTableGain}{$\uparrow$.004} & 66.5\rlap{\strideDelta{strideTableGain}{$\uparrow$3.5}} & 4.6\strideDelta{strideTableLoss}{$\uparrow$.8} & 0.605\strideDelta{strideTableLoss}{$\downarrow$.006} & 0.873\strideDelta{strideTableLoss}{$\downarrow$.027} \\
SATO~\cite{sato} & 0.0325\strideDelta{strideTableGain}{$\downarrow$.0049} & \underline{0.0098}\strideDelta{strideTableGain}{$\downarrow$.0177} & \underline{0.783}\strideDelta{strideTableGain}{$\uparrow$.212} & 0.902\strideDelta{strideTableTie}{0} & \underline{70.0}\rlap{\strideDelta{strideTableGain}{$\uparrow$7.0}} & \underline{2.7}\strideDelta{strideTableGain}{$\downarrow$1.1} & \underline{0.792}\strideDelta{strideTableGain}{$\uparrow$.181} & 0.890\strideDelta{strideTableLoss}{$\downarrow$.010} \\
\midrule
\strideband{1}{strideTableOurs}
\textbf{STRIDE (ours)} & \underline{0.0322}\strideDelta{strideTableGain}{$\downarrow$.0052} & \textbf{0.0074}\strideDelta{strideTableGain}{$\downarrow$.0201} & \textbf{0.835}\strideDelta{strideTableGain}{$\uparrow$.264} & \textbf{0.927}\strideDelta{strideTableGain}{$\uparrow$.025} & \textbf{72.0}\rlap{\strideDelta{strideTableGain}{$\uparrow$9.0}} & \textbf{2.5}\strideDelta{strideTableGain}{$\downarrow$1.3} & \textbf{0.836}\strideDelta{strideTableGain}{$\uparrow$.225} & \textbf{0.920}\strideDelta{strideTableGain}{$\uparrow$.020} \\
\bottomrule
\end{tabular*}

\end{table*}
\textbf{Backbone and data.} ECHO Lite~\cite{echo} generates robot-native G1 motion with a conditional UNet. The benchmark contains 10,792 training and 2,084 test motions, with rewordings from several sources and left/right edits. FRoM-W1~\cite{fromw1} provides a held-out perturbation source.

\textbf{Two comparison protocols.} The benchmark uses frozen-reference supervision on noised data, whereas the controlled comparison also anchors the original prediction on sampled states ($\lambda_o=0$ and $\lambda_o=2$, respectively). Table~\ref{tab:liteopd}(b) varies only the preserve target. Proposition~\ref{prop:stability} gives a sufficient condition at a fixed preserve state, with weights adjusted for minibatch composition.

\textbf{Joint data and training.} Of 313,972 command pairs, 309,914 preserve meaning and the rest change direction, speed or gait. We sample rewordings more often and give each edit type equal probability. Both models start from the same ECHO Lite weights and share the minibatches and loss coefficients. Same-input on-policy distillation (OPD) uses $T_p$ for rewordings and \method{} uses $\mathrm{sg}(F_o)$. We use the same validation rule for both models.

\textbf{Matched training budget.} Each run uses 50k updates at batch size 32 with trajectory weights $(\lambda_p,\lambda_e,\lambda_o)=(4,4,2)$. We select the checkpoint with the best combined retrieval and RepCos among those close to the best validation FID. This selection uses held-out validation motions throughout.

\textbf{Inference.} Table~\ref{tab:main} averages motion generation metrics over seeds 45, 46 and 47 using the released ten-step DPM-Solver~\cite{dpmsolver} and classifier-free guidance of 2.5~\cite{cfg}. Paired commands share the noise draws, except that WINRO selects from a common candidate pool. The complete models retain their learned registers at inference and take a single command.

\textbf{Core metrics.} In ECHO's released MoCLIP motion space, FID compares original-command generations with real motions. FID$_D$ compares original and reworded generations, and RepCos gives their paired cosine similarity. We average both measures across lexical, syntactic and narrative rewordings. R@3 tests whether the generated motion ranks its own caption among the top three of 32 candidates. Following behavioral tests of invariance and directional change~\cite{checklist}, we distinguish Edit Flip (activity shifts toward the requested side) from paired-side accuracy (both commands use the correct side).

\textbf{Video-language check.} Video-language models can assess motion from rendered clips~\cite{jivlmeval}. In our test, each model generates from the synonym rewrite. Qwen3.5-9B receives the original caption and two 64-frame clips: the dataset reference first, then the anonymous generation. Only a full match counts.

\textbf{Embodied evaluation.} ECHO's deployment client integrates velocities and smooths the reference before adding a static start. The GentleHumanoid-based tracker~\cite{gentlehumanoid} executes it in MuJoCo~\cite{mujoco}. We count a failure when root-height error is above 0.25\,m or torso tilt error is above $50^\circ$. Paraphrase-induced failure (PIF) averages each successfully tracked original's fraction of failed rewrites. xRepCos and xFlip measure motion consistency and edit response after tracking. xRTC compares the shape and distance of executed horizontal paths.

\textbf{Benchmark methods.} We compare published ECHO, adapted LwF, LEAF, MoCLIP and SATO, inference-only MultiAct and WINRO, and \method{}. WINRO selects noise for each caption from the same eight candidates.

\begin{figure*}[t]\centering
\includegraphics[width=.92\textwidth]{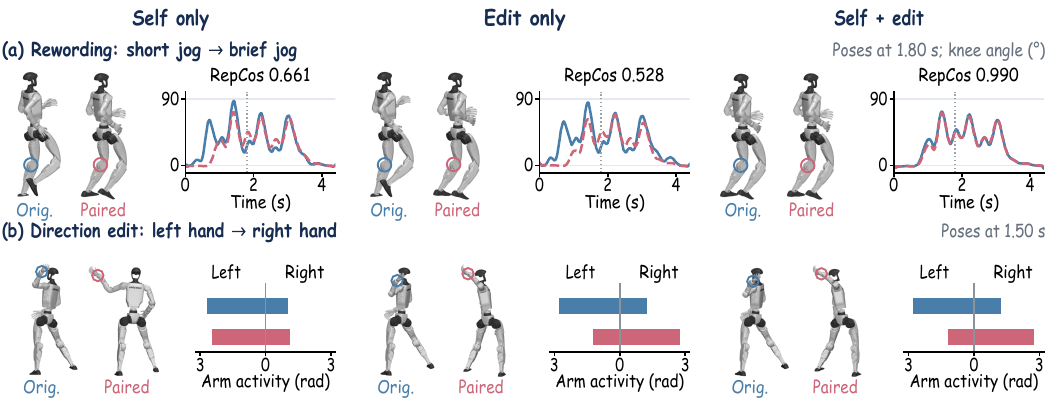}
\caption{\textbf{Joint anchoring retains both responses in these selected examples.} (a) Rewording. (b) Left-to-right edit. Poses accompany full-motion measurements. Blue marks the original command and rose the paired command. The jog illustrates a joint-model RepCos advantage of at least 0.08 over both single targets.}
\label{fig:ablation_summary}
\end{figure*}

\section{Selective Robustness: Results and Mechanism}
\label{sec:results}
\method{} improves wording consistency over SATO before and after tracking and retains a higher executed flip rate (Table~\ref{tab:main}). On synonym rewrites, the VLM scored 72.0\% of \method{} motions and 63.0\% of ECHO motions as full matches.

\begin{table}[!htb]\centering
\caption{\textbf{Trajectory targets and G1 execution.} (a) Shared-trainer ablation with a separate base. (b,c) Matched targets. (d) Tracking for (b). Deltas use the yellow row.}
\label{tab:ablation}\label{tab:liteopd}\tabsetup
\resizebox{\linewidth}{!}{\setlength{\tabcolsep}{2.5pt}\begin{tabular}{@{}lllll@{}}
\toprule
\strideband{1}{strideTableHead}
\bfseries Method  &  \bfseries FID $\downarrow$  &  \bfseries FID$_D$ $\downarrow$  &  \bfseries RepCos $\uparrow$  &  \bfseries Edit Flip $\uparrow$ \\
\midrule
\multicolumn{5}{@{}l}{\emph{(a) Trajectory terms: shared-trainer ablation}} \\
\strideband{1}{strideTableStripe}
No traj. terms & 0.0338 & 0.0081 & 0.823 & 0.906 \\
Self only & \underline{0.0325}\strideDelta{strideTableLoss}{$\uparrow$.0007} & \textbf{0.0062}\strideDelta{strideTableGain}{$\downarrow$.0013} & \textbf{0.860}\strideDelta{strideTableGain}{$\uparrow$.033} & 0.881\strideDelta{strideTableLoss}{$\downarrow$.039} \\
\strideband{1}{strideTableStripe}
Edit only & 0.0329\strideDelta{strideTableLoss}{$\uparrow$.0011} & 0.0083\strideDelta{strideTableLoss}{$\uparrow$.0008} & 0.819\strideDelta{strideTableLoss}{$\downarrow$.008} & \underline{0.917}\strideDelta{strideTableLoss}{$\downarrow$.003} \\
\strideband{1}{strideTableOurs}
Self + edit & \textbf{0.0318} & \underline{0.0075} & \underline{0.827} & \textbf{0.920} \\
\midrule
\multicolumn{5}{l}{\emph{(b) Joint multi-edit: preserve-target comparison}} \\
\strideband{1}{strideTableStripe}
Same-input OPD & \underline{0.0349}\strideDelta{strideTableLoss}{$\uparrow$.0016} & \underline{0.0118}\strideDelta{strideTableLoss}{$\uparrow$.0035} & \underline{0.783}\strideDelta{strideTableLoss}{$\downarrow$.032} & \textbf{0.918}\strideDelta{strideTableGain}{$\uparrow$.005} \\
\strideband{1}{strideTableOurs}
 \textbf{STRIDE} & \textbf{0.0333} & \textbf{0.0083} & \textbf{0.815} & \underline{0.913} \\
\midrule
\multicolumn{5}{l}{\emph{(c) Minimal core: preserve-target comparison}} \\
\strideband{1}{strideTableStripe}
Same-input OPD & \underline{0.0353}\strideDelta{strideTableLoss}{$\uparrow$.0030} & \underline{0.0196}\strideDelta{strideTableLoss}{$\uparrow$.0097} & \underline{0.694}\strideDelta{strideTableLoss}{$\downarrow$.093} & \underline{0.901}\strideDelta{strideTableLoss}{$\downarrow$.020} \\
\strideband{1}{strideTableOurs}
 \textbf{STRIDE} & \textbf{0.0323} & \textbf{0.0099} & \textbf{0.787} & \textbf{0.921} \\
\midrule
\multicolumn{5}{@{}l}{\emph{(d) G1 tracking of the models in (b)}} \\
\strideband{1}{strideTableHead}
& \multicolumn{2}{l}{\bfseries Both tracked (\%) $\uparrow$} & \bfseries xRTC $\uparrow$ & \bfseries xRepCos $\uparrow$ \\
\strideband{1}{strideTableStripe}
Same-input OPD & \multicolumn{2}{l}{\underline{91.67}\strideDelta{strideTableLoss}{$\downarrow$1.23}} & \underline{0.752}\strideDelta{strideTableLoss}{$\downarrow$.032} & \underline{0.778}\strideDelta{strideTableLoss}{$\downarrow$.036} \\
\strideband{1}{strideTableOurs}
\textbf{STRIDE} & \multicolumn{2}{l}{\textbf{92.90}} & \textbf{0.784} & \textbf{0.814} \\
\bottomrule
\end{tabular}}

\end{table}

\subsection{The preserve target improves wording consistency}
\label{sec:ablation}
\textbf{Matched target comparison.} With inputs and training budgets fixed, replacing the preserve target lowers FID$_D$ by 30.0\% and raises RepCos from 0.783 to 0.815 (Table~\ref{tab:liteopd}(b)). The gain carries over to G1 tracking, where both completion and executed path consistency improve (Table~\ref{tab:liteopd}(d)).

\textbf{What evidence alignment adds.} Even in the minimal core, the preserve target raises RepCos from 0.694 to 0.787. The joint recipe improves consistency further, but several components change together. A separate ablation isolates evidence alignment by fixing the recipe and omitting trajectory terms.

Removing evidence alignment in that ablation lowers RepCos from 0.823 to 0.803 and original R@3 from 0.801 to 0.786. The alignment term therefore helps even without rollout supervision.

\subsection{Edit anchoring retains direction responses}
Table~\ref{tab:ablation}(a) compares trajectory targets under one trainer. Self-only gives the highest RepCos. Adding edit anchoring retains a RepCos of 0.827 while raising paired-side accuracy from 54.9\% to 76.9\%. Conversely, adding self-anchoring to edit-only improves wording consistency.

\label{sec:retention}
\textbf{Dependence on the reference.} When the reference satisfies both commands, joint anchoring retains 95.1\% of correct pairs, compared with 67.6\% for self-only. The reference is a strong predictor of which edits survive training.

The joint model also corrects 58 of the 222 pairs on which the reference fails. The local bound constrains the size of an edit response, while the test measures whether the requested limb actually moves.

\subsection{One model responds to speed and action edits}
\label{sec:semantics}
To test responses beyond laterality, we evaluate changes in speed and action identity.
Table~\ref{tab:semantics} reports means over three shared generation seeds (42, 43 and 44) for 76 slow/fast pairs from 38 held-out families and 78 walk/run pairs from 39 families. The speed test requires the fast command to yield higher mean horizontal speed and both motions to be classified as walking. The walk/run test checks both labels with the same motion-only recognizer. Both tests also require the requested direction.

\begin{table}[!htb]\centering
\caption{\textbf{Joint-model speed and action responses.} Rates require both outputs to satisfy the criterion; $\Delta\nu$ is fast-minus-slow speed.}
\label{tab:semantics}\tabsetup\resizebox{\linewidth}{!}{\setlength{\tabcolsep}{3pt}\begin{tabular}{@{}lrrr@{}}
\toprule
\strideband{1}{strideTableHead}
\bfseries Method & \bfseries Speed + walk (\%) $\uparrow$ & \bfseries Walk/run (\%) $\uparrow$ & \bfseries $\Delta\nu$ (m/s) \\
\midrule
\strideband{1}{strideTableStripe}
ECHO Lite & 95.6 & 90.6 & 0.359 \\
Joint OPD & \underline{96.9} & \textbf{95.3} & 0.349 \\
\strideband{1}{strideTableOurs}
 \textbf{Joint STRIDE} & \textbf{97.8} & \underline{94.9} & 0.332 \\
\bottomrule
\end{tabular}}
\end{table}

The joint \method{} checkpoint satisfies the speed-and-walking criterion on 97.8\% of trials and both walk/run labels on 94.9\% (Tab.~\ref{tab:semantics}).

Figure~\ref{fig:ablation_summary} compares the underlying motions. Joint anchoring gives the highest RepCos for the illustrated jog. Only the edit-anchored models switch arms when the prompt changes side. Table~\ref{tab:semantics} tests whether the joint model also follows changes in speed and gait.

\subsection{Generalization across backbones, corpora and perturbations}
\label{sec:full}\label{sec:bones}\label{sec:textperturbations}
On ECHO Full, \method{} raises RepCos from 0.654 to 0.782 (Table~\ref{tab:fullecho}). On the BONES-SEED dataset from Bones Studio, only \method{} receives target-corpus fine-tuning. Both Lite models use the same corpus-adapted evaluator.

\begin{table}[!htb]\centering
\caption{\textbf{Backbone scaling and corpus adaptation.} Methods share an evaluator and sampler within each block.}
\label{tab:fullecho}\label{tab:boneslite}\tabsetup
\begin{tabular*}{\linewidth}{@{\extracolsep{\fill}}lrrrr@{}}
\toprule
\strideband{1}{strideTableHead}
\bfseries Method  &  \bfseries FID $\downarrow$  &  \bfseries FID$_D$ $\downarrow$  &  \bfseries RepCos $\uparrow$  &  \bfseries R@3 $\uparrow$ \\
\midrule
\multicolumn{5}{@{}l}{\emph{(a) Full ECHO backbone}} \\
\strideband{1}{strideTableStripe}
ECHO Full & \textbf{0.029} & \underline{0.006} & \underline{0.654} & \underline{0.688} \\
\strideband{1}{strideTableOurs}
 \textbf{STRIDE} & \underline{0.031} & \textbf{0.003} & \textbf{0.782} & \textbf{0.781} \\
\midrule
\multicolumn{5}{@{}l}{\emph{(b) BONES-SEED adaptation}} \\
\strideband{1}{strideTableStripe}
ECHO Lite & \underline{0.272} & \underline{0.042} & \underline{0.595} & \underline{0.584} \\
\strideband{1}{strideTableOurs}
 \textbf{STRIDE} & \textbf{0.192} & \textbf{0.027} & \textbf{0.773} & \textbf{0.680} \\
\bottomrule
\end{tabular*}

\end{table}

Across eight text perturbations, Full \method{} improves RepCos and fixed-caption retrieval in every family, with higher clean FID in seven. The tests include character triplication and intratoken spacing, both absent from the inspected training and validation texts.

\subsection{From generated trajectories to G1 execution}
\label{sec:tracking}

\begin{figure}[!htbp]\centering
\includegraphics[width=\columnwidth]{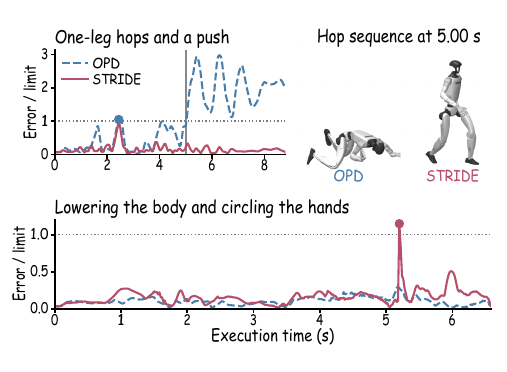}
\caption{\textbf{G1 tracking after rewording in MuJoCo.} Dots mark first limit crossings. At 5.00\,s on the hop rewrite, pelvis height is 0.37\,m for fallen OPD and 0.74\,m for upright \method{}.}
\label{fig:matchedtracking}
\end{figure}

Figure~\ref{fig:matchedtracking} shows one failure for each model. OPD falls on the hop rewrite. \method{} crosses the $50^\circ$ tilt limit at 5.20\,s while lowering its torso and circling its hands.

\method{} tracks more original commands than OPD, so their conditional failure rates use different denominators. On the 368 LLM originals that both models track, 12 rewrites fail only with OPD and four only with \method{}.

On the shared human-caption originals, five rewrites fail exclusively for each model. Averaging the failed-rewrite fraction per successful original gives 2.34\% for \method{} (761 originals) and 2.22\% for OPD (750).

\begin{figure}[!htbp]\centering
\includegraphics[width=\columnwidth]{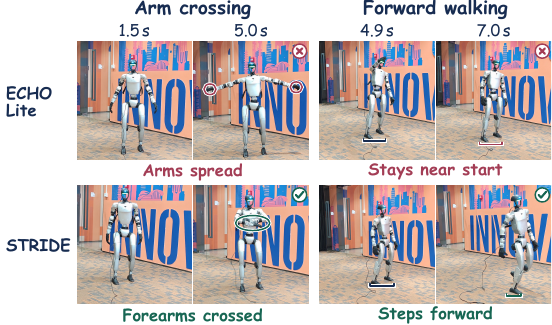}
\caption{\textbf{Two rewordings on a physical G1.} Only \method{} crosses its forearms on the first command and walks ahead on the second.}
\label{fig:hardware-execution}
\end{figure}

\textbf{Physical G1 execution.} For \emph{the individual has arms intersecting}, ECHO Lite spreads its arms and \method{} crosses them. For \emph{human is treading forward}, ECHO Lite moves its limbs near the starting point while \method{} walks ahead (Figure~\ref{fig:hardware-execution}).

In the walking references, the rewrite reduces net horizontal root displacement from 2.54 to 0.06\,m for ECHO Lite and from 2.33 to 2.27\,m for \method{} at the same seed.

\subsection{Motion-conditioned attribution analysis}
\label{sec:attr_results}
Centering and prediction separate matched from unrelated captions even when uncentered pooling gives both a high cosine. Across 2,084 Full \method{} pairs, Grad$\times$Input (GI) assigns 70.5\% of mass to lexical words and 1.7\% to padding, compared with 8.6\% and 58.5\% for attention. Lexical words also retain most of the normalized pooling weight across the noise levels tested.

To test whether token rankings predict output sensitivity~\cite{saliencysanity}, we attenuate the top-ranked embeddings. At $t=500$, scaling the top fifth by 0.8 shifts the Full \method{} prediction by 0.0086 RMSE under GI and 0.0053 under attention. Over full rollouts, attenuating the top GI tokens gives the largest mean change. Leaving them intact and attenuating the rest gives the smallest (Table~\ref{tab:attr_rollout}(a)).

On frozen Full \method{}, GI best separates paired from unrelated captions and gives the highest effective rank. Changing only the pooling rule in Lite training also favors GI on RepCos, FID$_D$ and retrieval (Table~\ref{tab:attr_rollout}(b)).

Both self-only and joint training register a direction edit at the input. Replacing a paraphrase with a left/right edit raises GI mass on side words from 4.15\% to 5.57\% for self-only and from 3.81\% to 5.42\% for joint anchoring. Their paired-side accuracy nevertheless differs, as the target ablation shows.

\begin{table}[!htb]\centering
\caption{\textbf{Token ranking and pooling.} (a) Mean cosine drop from clean motion on 64 families. (b) Original ($o$) and rewritten ($r$) retrieval.}
\label{tab:attr_rollout}\tabsetup\begingroup\setlength{\tabcolsep}{1.3pt}
\begin{tabular*}{\linewidth}{@{\extracolsep{\fill}}lrrrrrr@{}}
\toprule
\strideband{2}{strideTableHead}
\emph{(a)} & \multicolumn{3}{c}{Attenuate top 20\% $\uparrow$} & \multicolumn{3}{c}{Retain top 30\% $\downarrow$} \\
\cmidrule(lr){2-4}\cmidrule(lr){5-7}
Model & GI & Attn. & Random & GI & Attn. & Random \\
\midrule
\strideband{1}{strideTableStripe}
ECHO Full & \textbf{0.1343} & \underline{0.0846} & 0.0707 & \textbf{0.1045} & \underline{0.1204} & 0.1216 \\
\strideband{1}{strideTableOurs}
STRIDE Full & \textbf{0.0249} & \underline{0.0150} & 0.0137 & \textbf{0.0122} & 0.0353 & \underline{0.0307} \\
\strideband{1}{strideTableStripe}
ECHO Lite & \textbf{0.0976} & \underline{0.0626} & 0.0516 & \textbf{0.0696} & \underline{0.0918} & 0.0964 \\
\strideband{1}{strideTableOurs}
STRIDE Lite & \textbf{0.0092} & 0.0044 & \underline{0.0062} & \textbf{0.0078} & \underline{0.0125} & 0.0140 \\
\bottomrule\end{tabular*}\par\smallskip
\begin{tabular*}{\linewidth}{@{\extracolsep{\fill}}lrrrrrr@{}}
\strideband{2}{strideTableHead}
\emph{(b)} & \multicolumn{2}{c}{Frozen Full} & \multicolumn{4}{c}{Matched Lite training (10k)} \\
\cmidrule(lr){2-3}\cmidrule(lr){4-7}
Pool & Margin & Rank & RepCos & FID$_D$ & R@3$_o$ & R@3$_r$ \\
\midrule
\strideband{1}{strideTableStripe}
Uniform & 0.4167 & 42.32 & 0.8046 & 0.00879 & 0.3914 & 0.3175 \\
Attention & \underline{0.4198} & \underline{43.00} & \underline{0.8098} & \underline{0.00876} & \underline{0.4007} & \underline{0.3217} \\
\strideband{1}{strideTableOurs}
GI & \textbf{0.4841} & \textbf{49.67} & \textbf{0.8130} & \textbf{0.00841} & \textbf{0.4065} & \textbf{0.3223} \\
\bottomrule
\end{tabular*}\endgroup

\end{table}

\textbf{When wording changes the action.} Relative to ECHO Full, \method{} gains 0.252 RepCos on LLM rewrites and 0.140 on rule variants, but loses 0.078 on human captions. Human annotators can describe different phases of the same motion, making global alignment inappropriate for those pairs.

Speed edits expose a different limit. Training only on left/right edits narrows the fast--slow gap from 0.359 to 0.226\,m/s, while multi-edit training recovers a gap of 0.332\,m/s (Table~\ref{tab:semantics}). With an expanded edit vocabulary, residual errors cluster near the boundary between fast walking and running.

\section{Discussion and Conclusion}
Attribution identifies words that affect the predicted motion. The rollout target controls which differences persist. Self-anchoring brings paraphrase predictions closer. Edit anchoring retains a response when the reference separates the commands and the anchor errors are small (Proposition~\ref{prop:edit}). Under matched training, paired MuJoCo completion rises from 91.7\% to 92.9\% (Tab.~\ref{tab:liteopd}(d)). On the physical G1, the clearest difference was forward progress under the walking rewrite.

In that example, attenuating \emph{treading} at successive solver steps would test when the verb affects final displacement. Attributing each limb and phase separately could guide local anchors. Tracker feedback could supply targets for edits the reference misses.\label{stride:main-end}

{\def\bibfont{\footnotesize}\setlength{\bibsep}{2pt plus 0.5pt minus 0.5pt}\bibliographystyle{ieeenat_fullname}\bibliography{refs}}
\clearpage
\finalhyphendemerits=100000
\setcounter{section}{0}
\setcounter{table}{0}
\setcounter{figure}{0}
\setcounter{equation}{0}
\setcounter{algorithm}{0}
\renewcommand{\thesection}{S\arabic{section}}
\renewcommand{\thetable}{S\arabic{table}}
\renewcommand{\thefigure}{S\arabic{figure}}
\renewcommand{\theequation}{S\arabic{equation}}
\renewcommand{\thealgorithm}{S\arabic{algorithm}}
\renewcommand{\theHsection}{supp.S\arabic{section}}
\renewcommand{\theHtable}{supp.S\arabic{table}}
\renewcommand{\theHfigure}{supp.S\arabic{figure}}
\renewcommand{\theHequation}{supp.S\arabic{equation}}
\renewcommand{\theHalgorithm}{supp.S\arabic{algorithm}}
\setcounter{dbltopnumber}{2}
\setcounter{topnumber}{2}
\setcounter{bottomnumber}{1}
\setcounter{totalnumber}{3}
\renewcommand{\dbltopfraction}{0.9}
\renewcommand{\dblfloatpagefraction}{0.82}
\setlength{\textfloatsep}{8pt plus 1pt minus 2pt}
\setlength{\dbltextfloatsep}{8pt plus 1pt minus 2pt}
\setlength{\dblfloatsep}{8pt plus 2pt minus 2pt}
\makeatletter
\setlength{\@dblfptop}{0pt}
\setlength{\@dblfpsep}{8pt}
\setlength{\@dblfpbot}{0pt plus 1fil}
\setlength{\@fptop}{0pt}
\setlength{\@fpsep}{8pt}
\setlength{\@fpbot}{0pt plus 1fil}
\makeatother
\maketitlesupplementary
\raggedbottom
\setcounter{dbltopnumber}{1}
\renewcommand{\dbltopfraction}{0.9}
\renewcommand{\dblfloatpagefraction}{0.82}
\extrafloats{80}
\newcommand{\appdisplaytext}[1]{\parbox{\linewidth}{\normalfont\normalsize #1}\par\smallskip}
\newenvironment{appdisplay}
  {\begin{table*}[t]\centering}
  {\end{table*}}
\newenvironment{ceblock}
  {\begin{table*}[t]\centering}
  {\end{table*}}
\captionsetup{skip=3pt,hypcap=false}
\newcommand{\ceci}[3]{#1}
\newcommand{\cetab}{\tabsetup}
\newcommand{\ceOriginalFIDDelta}{+0.001820}
\newcommand{\ceGIEMAMarginDelta}{+0.015854}
\newcommand{\ceGIEMAMarginLow}{-0.002373}
\newcommand{\ceGIEMAMarginHigh}{+0.034111}
\newcommand{\ceAttentionMarginDelta}{-0.162790}
\newcommand{\ceAttentionMarginLow}{-0.192597}
\newcommand{\ceAttentionMarginHigh}{-0.132871}
\newcommand{\ceUniformMarginDelta}{-0.058841}
\newcommand{\ceUniformMarginLow}{-0.076615}
\newcommand{\ceUniformMarginHigh}{-0.041570}

\section{Results behind the main comparisons}
Tables, figures and equations without an S prefix refer to the main paper. Together with this document, the supplementary archive provides \texttt{real\_\allowbreak robot\_\allowbreak trials.mp4} (34\,s, 1080p, 30\,fps side-by-side physical Unitree G1 executions across five reworded tasks at $1\times$ speed; Section~\ref{app:sim}, Table~\ref{tab:robot_video}) and \texttt{code/} (training recipes, attribution verification and measurement reproduction). Sections~\ref{app:implementation}--\ref{supp:attribution} cover the training update and proofs, benchmark curation and evaluators, controlled target and edit comparisons, cross-corpus and perturbation tests, execution, and attribution.

\textbf{Deployment invariance and protocol design.} At inference, \method{} discards all auxiliary alignment heads ($P,Q,\mu$), teacher copies and feature perturbations, retaining only the generator and lightweight learned padding registers ($<0.02\%$ of model parameters). Deployment keeps the native ten-step DPM-Solver (CFG 2.5) and takes a single command without relation labels. Across training protocols, Table~1 sets $\lambda_o=0$ because noised-data fidelity and preserve reference anchoring already tether $f_\theta$ to $f_{\mathrm{ref}}$, whereas Table~2(b,c) activates sampled-state original anchoring ($\lambda_o=2$) to isolate the trajectory preserve target ($\mathrm{sg}(F_o)$ versus $T_p$) under a shared stability anchor. Under the $80\%$ preserve mixture in Section~\ref{app:joint}, these weights satisfy Proposition~3.2 without modifying inference (Section~\ref{app:theory}).

\textbf{Why detached attribution pooling trains text grounding.} Under native cross-attention, padding positions absorb the majority of attention weight while lexical words receive only a small share, whereas reconstruction Grad$\times$Input concentrates sensitivity on lexical words and largely ignores padding (Section~\ref{app:attr_sweeps}). Detaching the pooling weights $\alpha_i$ in Eq.~\ref{eq:pool} avoids unstable second-order Hessian paths through the UNet denoiser while routing gradients from the centered predictor loss $\mathcal L_{\mathrm{align}}$ (Eq.~\ref{eq:align}) directly into the trainable CLIP tower, text Transformer, projection $P$ and predictor $Q$. In parallel, learned registers replace caption-dependent padding features after EOS and train end-to-end under the base denoising and trajectory objectives.

\textbf{Reproducing reported measurements and attribution checks.} The \texttt{code/} directory provides two standalone CPU verifiers alongside the training and evaluation scripts (\texttt{train.py}, \texttt{train\_\allowbreak joint.py} and \texttt{train\_\allowbreak pooling.py}, plus \texttt{eval\_\allowbreak generate.py}, \texttt{eval\_\allowbreak metrics.py} and \texttt{eval\_\allowbreak edit.py}). Running \texttt{check\_\allowbreak attribution.py} verifies masked reconstruction scoring (Eq.~1), coordinatewise Grad$\times$Input magnitudes and relative-softmax content pooling with uniform fallback (Eq.~\ref{eq:pool}). Running \texttt{summarize\_\allowbreak measurements.py} recomputes the three-seed generation means and paired video verdicts (Tables~\ref{supp:seedrows}--\ref{supp:judges}), the validation FID checkpoint selections (Table~\ref{supp:selection}), the joint and minimal-core target-isolation gains (Table~2(b,c)), the paired pooling intervals (Table~\ref{tab:lite_pool_training}), and the MuJoCo completion, PIF and discordant-failure counts in Table~2(d), matching every reported value.

\textbf{Matched target isolation and paired evaluation.} Table~1 compares methods on ECHO Lite across seeds 45--47 under shared initial and per-step solver noise, separating wording sensitivity from within-command sampling diversity (\emph{Floor} in Table~\ref{tab:main:detailed}). Across all three sampling seeds, \method{} preserves tight run-to-run stability in both RepCos and $\mathrm{FID}_D$ (Table~\ref{supp:seedrows}). Table~2(b) isolates the preserve target at seed~42 with identical multi-edit data, edit and original anchors, and a 50k-update budget: matching the student's own original-command prediction $\mathrm{sg}(F_o)$ instead of the frozen reference $T_p$ on rewordings directly lowers the wording-induced distribution gap.

\textbf{Which checkpoint enters each comparison?} Every run supplies one candidate per 10,000 updates. We keep checkpoints within 10\% of that run's lowest validation FID, then select the largest sum of its retrieval and representation scores. Table~\ref{supp:selection} shows how many candidates pass and which one enters the test comparison.
\begin{appdisplay}
\captionof{table}{\textbf{Validation checkpoint selection.} The FID limit is $1.10$ times the minimum validation FID in each run. All five runs supply five candidates. Section~\ref{app:bones} gives the BONES-SEED split and scoring rule.}\label{supp:selection}
\tabsetup\begin{tabular*}{.94\linewidth}{@{\extracolsep{\fill}}llrrr@{}}
\toprule\strideband{1}{strideTableHead}
\textbf{Training set} & \textbf{Method} & \textbf{FID limit} & \textbf{Eligible} & \textbf{Selected update}\\\midrule
Joint multi-edit & Same-input OPD & 0.14841 & 5 & 10,000\\
\strideband{1}{strideTableStripe}
Joint multi-edit & STRIDE & 0.14744 & 5 & 10,000\\
Minimal core & Same-input OPD & 0.15152 & 5 & 10,000\\
\strideband{1}{strideTableStripe}
Minimal core & STRIDE & 0.14767 & 5 & 10,000\\
BONES-SEED & STRIDE & 0.35758 & 2 & 40,000\\
\bottomrule
\end{tabular*}
\end{appdisplay}

\textbf{Sampling variation in Table~1.} Across all three generation seeds, \method{} achieves the highest RepCos, highest Edit Flip and lowest FID$_D$, while LwF keeps the lowest original-command FID by tethering directly to the frozen teacher (Table~\ref{supp:seedrows}). Each seed evaluates all 2,084 test motions and 840 opposite-side edit pairs.
\begin{appdisplay}
\captionof{table}{\textbf{Generation results by sampling seed.} Each seed uses the same 2,084 test commands and 840 edit pairs; bold and underline mark the best and second-best displayed scores per column.}\label{supp:seedrows}
\tabsetup\begin{tabular*}{\linewidth}{@{\extracolsep{\fill}}l*{12}{r}@{}}
\toprule\strideband{2}{strideTableHead}
 & \multicolumn{3}{c}{FID $\downarrow$} & \multicolumn{3}{c}{FID$_D$ $\downarrow$} & \multicolumn{3}{c}{RepCos $\uparrow$} & \multicolumn{3}{c}{Edit Flip $\uparrow$}\\
\cmidrule(lr){2-4}\cmidrule(lr){5-7}\cmidrule(lr){8-10}\cmidrule(lr){11-13}
\textbf{Method} & 45 & 46 & 47 & 45 & 46 & 47 & 45 & 46 & 47 & 45 & 46 & 47\\\midrule
\strideband{1}{strideTableStripe}
ECHO & 0.0373 & 0.0370 & 0.0380 & 0.0279 & 0.0265 & 0.0281 & 0.573 & 0.572 & 0.567 & \underline{0.908} & 0.901 & 0.895 \\
MultiAct & 0.0387 & 0.0384 & 0.0395 & 0.0276 & 0.0265 & 0.0281 & 0.570 & 0.570 & 0.566 & 0.904 & 0.902 & 0.899 \\
WINRO & 0.0377 & 0.0397 & 0.0376 & 0.0273 & 0.0276 & 0.0285 & 0.515 & 0.507 & 0.510 & 0.902 & 0.890 & 0.899 \\
LwF & \textbf{0.0304} & \textbf{0.0309} & \textbf{0.0310} & 0.0270 & 0.0263 & 0.0264 & 0.573 & 0.571 & 0.572 & 0.898 & 0.887 & 0.892 \\
LEAF & 0.0375 & 0.0365 & 0.0377 & 0.0270 & 0.0259 & 0.0271 & 0.578 & 0.582 & 0.576 & 0.905 & 0.906 & \underline{0.907} \\
MoCLIP & 0.0388 & 0.0387 & 0.0391 & 0.0284 & 0.0279 & 0.0277 & 0.576 & 0.577 & 0.576 & 0.904 & \underline{0.913} & 0.902 \\
SATO & 0.0329 & 0.0322 & \underline{0.0323} & \underline{0.0097} & \underline{0.0100} & \underline{0.0098} & \underline{0.783} & \underline{0.782} & \underline{0.785} & 0.893 & 0.907 & 0.905 \\
\midrule
\strideband{1}{strideTableOurs}
\textbf{STRIDE} & \underline{0.0318} & \underline{0.0320} & 0.0328 & \textbf{0.0072} & \textbf{0.0074} & \textbf{0.0075} & \textbf{0.836} & \textbf{0.835} & \textbf{0.834} & \textbf{0.917} & \textbf{0.935} & \textbf{0.929} \\
\bottomrule
\end{tabular*}
\end{appdisplay}

\textbf{Which videos changed the verdict?} On the 200 shared synonym rewrites judged from rendered clips at seed 42, \method{} reaches the highest full-match count, with paired discordant flips favoring \method{} over every baseline including SATO (Table~\ref{supp:judges}). Visual judgment thus corroborates the embedding-space consistency gains.
\begin{appdisplay}
\captionof{table}{\textbf{Video judgments on 200 shared synonym rewrites.} Five-way Qwen3.5-9B verdicts and discordant full matches vs.\ STRIDE.}\label{supp:judges}
\tabsetup\begin{tabular*}{.94\linewidth}{@{\extracolsep{\fill}}lrrrrrrr@{}}
\toprule\strideband{2}{strideTableHead}
 & \multicolumn{5}{c}{Verdict on the generated clip} & \multicolumn{2}{c}{Paired with STRIDE}\\
\cmidrule(lr){2-6}\cmidrule(lr){7-8}
\textbf{Method} & Full $\uparrow$ & Partial & Mismatch & Uncertain & Error & STRIDE only & Other only\\\midrule
\strideband{1}{strideTableStripe}
ECHO & 126 & 16 & 55 & 2 & 1 & 31 & 13 \\
MultiAct & 127 & 14 & 55 & 3 & 1 & 35 & 18 \\
WINRO & 129 & 16 & 51 & 3 & 1 & 37 & 22 \\
LwF & 128 & 12 & 57 & 2 & 1 & 28 & 12 \\
LEAF & 131 & 16 & 51 & 1 & 1 & 32 & 19 \\
MoCLIP & 133 & 12 & 54 & 0 & 1 & 31 & 20 \\
SATO & \underline{140} & 19 & 38 & 2 & 1 & 20 & 16 \\
\midrule
\strideband{1}{strideTableOurs}
\textbf{STRIDE} & \textbf{144} & 12 & 41 & 2 & 1 & -- & -- \\
\bottomrule
\end{tabular*}
\end{appdisplay}

\textbf{From token sensitivity to motion.} On frozen Full \method{}, preserving the top-ranked tokens while attenuating the rest retains much closer motion similarity under Grad$\times$Input ranking than under cross-attention ranking (Figure~\ref{fig:attr_intervene}). In matched Lite training, Grad$\times$Input pooling likewise achieves the best consistency, distribution distance and retrieval among uniform, attention and attribution pooling, with a clear paired gain over uniform pooling (Table~\ref{tab:lite_pool_training}).
\begin{table}[ht]
\centering
\caption{\textbf{Token pooling during Lite training.} Matched 10k-update runs reporting R@3 on original and rewritten captions.}
\label{tab:lite_pool_training}
{\cetab\small\renewcommand{\arraystretch}{1.15}\begin{tabular}{@{}lrrrr@{}}
\toprule
\strideband{1}{strideTableHead}
Pooling & RepCos$\uparrow$ & FID$_D\downarrow$ & R@3$_{\rm orig}\uparrow$ & R@3$_{\rm rew}\uparrow$ \\
\midrule
\strideband{1}{strideTableStripe}
Uniform & 0.8046 & 0.00879 & 0.3914 & 0.3175 \\
Attention & \underline{0.8098} & \underline{0.00876} & \underline{0.4007} & \underline{0.3217} \\
\strideband{1}{strideTableOurs}
Grad$\times$Input & \textbf{0.8130} & \textbf{0.00841} & \textbf{0.4065} & \textbf{0.3223} \\
\bottomrule
\end{tabular}
}
\end{table}

\textbf{G1 completion, laterality and physical execution.} Under equal per-original weighting across all 800 original commands in Table~2(d), \method{} completes more paired LLM and human rewrites than same-input OPD because its trajectory anchor prevents reworded references from drifting into unstable poses. In the shared-trainer ablation (Table~2(a), Table~\ref{tab:pairedside}), adding edit anchoring recovers paired left/right correctness that self-anchoring alone suppresses, a gap that single-direction Edit Flip does not expose. On physical Unitree G1 execution, this consistency carries over to hardware: ECHO Lite's generated walking reference collapses to near-zero forward travel under rewording, whereas \method{} maintains consistent forward displacement across both wordings (Section~\ref{app:sim}, Table~\ref{tab:robot_video}).

\section{Training targets and\texorpdfstring{\protect\\}{ }reconstruction evidence}
\label{app:implementation}
This section details the sampled-state rollout diagnostic, the complete \method{} training update in Algorithm~\ref{alg:stride}, and each term in the denoising, evidence-alignment and text-interface objectives. Section~\ref{app:theory} derives the clean-motion conversion and proves Propositions~3.1 and~3.2, and Section~\ref{app:hparams} records all loss weights, optimizer settings and single-GPU compute costs.

\textbf{Sampled-state diagnostic.} We compare predictions at matched timesteps on noised data and sampler-visited states across 256 narrative rewordings. Both SATO and the no-trajectory variant develop much larger prediction gaps on sampler states than on noised data, whereas \method{} narrows that gap along the rollout (Figure~3(a)). Confidence intervals use 4,000 paired bootstrap resamples of canonical motion families with mirrors grouped.

Algorithm~\ref{alg:stride} scores both captions on a shared noised motion and supervises their predictions at one rollout state.
\begin{algorithm}[tb]
\caption{One \method{} training update}\label{alg:stride}
\footnotesize
\begin{algorithmic}[1]
\algtext*{EndIf}\algtext*{EndFor}
\Require Minibatch $\mathcal B=\{(x_0^i,M^i,c^i,c^{\prime i},r_i)\}_{i=1}^{B}$
\Require $f_\theta$, frozen $f_{\mathrm{ref}}$, $\mathcal R_\theta$; time/noise laws $p_t,p_z$
\State $k\sim\mathrm{Unif}\{1,\ldots,K-1\}$
\For{$i=1,\ldots,B$}
  \State $\epsilon_i\sim\mathcal N(0,I),\quad t_i\sim p_t,\quad z_i\sim p_z$
  \State $x_{t_i}^{i}\gets\alpha_{t_i}x_0^i+\sigma_{t_i}\epsilon_i$
  \State $s_k^i\gets\mathrm{sg}[\mathcal R_\theta^{(k)}(c^{\prime i},z_i)]$
  \State $F_{o,i}\gets f_\theta(s_k^i,t_k,c^i),\; F_{p,i}\gets f_\theta(s_k^i,t_k,c^{\prime i})$
  \State $T_{o,i}\gets f_{\mathrm{ref}}(s_k^i,t_k,c^i)$
  \If{$r_i=\pre$}
    \State $y_i\gets\mathrm{sg}(F_{o,i})$
  \Else
    \State $T_{p,i}\gets f_{\mathrm{ref}}(s_k^i,t_k,c^{\prime i}),\; y_i\gets T_{p,i}$
  \EndIf
  \State $\ell_{r,i}\gets\|F_{p,i}-y_i\|_{M^i}^2,\;\ell_{o,i}\gets\|F_{o,i}-T_{o,i}\|_{M^i}^2$
\EndFor
\State $\mathcal L_{\mathrm{base}}\gets\mathcal L_{\mathrm{base}}(\mathcal B,\{x_{t_i}^{i},t_i\}_{i=1}^{B})$
\State $\mathcal L\gets\mathcal L_{\mathrm{base}}+\lambda_p\langle\ell_{r,i}\rangle_{\pre}+\lambda_e\langle\ell_{r,i}\rangle_{\edit}+\lambda_o\langle\ell_{o,i}\rangle$
\State $\theta\gets\mathsf{OptimizerStep}(\theta,\nabla_\theta\mathcal L)$
\end{algorithmic}
\smallskip
$\mathsf{OptimizerStep}$ updates $\theta$ and the optimizer state.
\end{algorithm}

\textbf{Notation and reductions.} For a residual $U$, we define the masked absolute error with $Z$ from Eq.~1 as
\begin{equation*}
\|U\|_{1,M}=\frac{1}{Z}\sum_{\ell=1}^{L}\sum_{d=1}^{D}M_\ell|U_{\ell d}|.
\end{equation*}
The base losses use one data state $(x_t,t)$ per pair, while trajectory losses use the shared sampler state $(s_k,t_k)$. Angular brackets denote means within a worker's minibatch; a relation subscript selects examples of that relation, and an empty group contributes zero. Distributed training averages the resulting gradients across workers, and $\varnothing$ denotes unconditional prediction.

\textbf{Base denoising and fidelity.} Define per-example data and reference errors
\begin{equation*}
\begin{aligned}
\ell_{\mathrm{den}}(c)&=\|f_\theta(x_t,t,c)-x_0\|_M^2,\\
\ell_{\mathrm{fid}}(c)&=\|f_\theta(x_t,t,c)-f_{\mathrm{ref}}(x_t,t,c)\|_{1,M}.
\end{aligned}
\end{equation*}
Both errors are functions of the conditioning command at the fixed data state. With fidelity weight $\lambda_f$ and unconditional weight $\lambda_u$, we define
\begin{equation*}
\begin{aligned}
\mathcal L_{\mathrm{unc}}&=\langle \ell_{\mathrm{den}}(\varnothing)+\lambda_f \ell_{\mathrm{fid}}(\varnothing)\rangle,\\
\mathcal L_{\mathrm{den}}&=\langle \ell_{\mathrm{den}}(c)+\lambda_f \ell_{\mathrm{fid}}(c)\rangle
+\langle \ell_{\mathrm{den}}(c')\rangle_{\pre}
+\lambda_u\mathcal L_{\mathrm{unc}}.
\end{aligned}
\end{equation*}
The original real motion supervises the original command and rewordings. Edited commands receive their trajectory targets from Eq.~4. The minimal-core configuration sets $\mathcal L_{\mathrm{base}}=\mathcal L_{\mathrm{den}}$ without auxiliary interface terms.

\textbf{Evidence alignment and its gradient path.}
In the Full model, $m_i$ denotes the reconstruction sensitivity before detachment, $a_i=\mathrm{sg}(m_i)$ the same scalar treated as constant during backpropagation, and $\alpha_i$ the pooling weight. With conditional clean-motion prediction, exact valid-frame normalization and shared $(x_0,x_t,t)$ for both wordings,
\begin{equation*}
\begin{aligned}
q_{\mathrm{rec}}(c)&=-\|f_\theta(x_t,t,c)-x_0\|_M^2,\\
g_i&=\nabla_{e_i}q_{\mathrm{rec}}(c),\qquad
m_i=\sum_d|e_{id}g_{id}|.
\end{aligned}
\end{equation*}
Let $I_c$ contain the non-special BPE tokens between BOS and EOS, including punctuation, and $\bar m$ be their mean magnitude. With $\tau_a=1$ and $\varepsilon_a=10^{-12}$,
\begin{equation}
\begin{aligned}
\alpha_i&=\mathrm{sg}\!\left[\operatorname{softmax}_{i\in I_c}\!\left(\frac{m_i}{\tau_a\max(\bar m,\varepsilon_a)}\right)\right],\\
u&=\sum_{i\in I_c}\alpha_iP(H_i),\qquad \sum_{i\in I_c}\alpha_i=1.
\end{aligned}
\label{eq:pool}
\end{equation}
With running center $\mu$, predictor $Q$ and centered vector $v=u-\mu$, the preserve alignment objective is
\begin{equation}
\begin{split}
\mathcal L_{\mathrm{align}}=-\tfrac12\big\langle
&\cos(Q(v),\mathrm{sg}(v'))\\
+&\cos(Q(v'),\mathrm{sg}(v))\big\rangle_{\pre}.
\end{split}
\label{eq:align}
\end{equation}
This aligns predicted, centered representations; token-attribution distributions are not matched directly. The input embedding $e_i\in\mathbb R^{512}$ precedes CLIP positional encoding, and its magnitude $m_i$ has a precise local interpretation. Under a coordinatewise relative perturbation $e_i+\delta(e_i\odot\xi)$,
\begin{equation*}
\begin{aligned}
\left.\frac{d q_{\mathrm{rec}}}{d\delta}\right|_{\delta=0}
  &=\langle e_i\odot g_i,\xi\rangle,\\
\max_{\|\xi\|_\infty\le1}|\langle e_i\odot g_i,\xi\rangle|
  &=\|e_i\odot g_i\|_1=m_i.
\end{aligned}
\end{equation*}
Thus $m_i$ measures worst-case linearized sensitivity over coordinatewise relative changes. Coherent scaling of the whole token instead gives $|g_i^\top e_i|\le m_i$, because opposite-sign coordinates can cancel before taking the absolute value; neither score is a finite token-deletion effect. Relative softmax uses all non-special BPE positions between BOS and EOS, including punctuation. Dividing by $\bar m$ removes a common positive score scale while the mean stays above the numerical floor, though weights still vary across timesteps; all-zero magnitudes fall back to uniform weights on this support.

The implementation stops gradients through $m_i$ and hence $\alpha_i$. The trainable feature $H_i\in\mathbb R^{256}$ is ECHO's output after CLIP, text projection, text Transformer and text normalization. Because the weights sum to one, applying the affine head $P(H_i)=WH_i+b$ before or after pooling is equivalent:
\begin{equation*}
u=\sum_{i\in I_c}\alpha_i(WH_i+b)=W\sum_{i\in I_c}\alpha_iH_i+b.
\end{equation*}
The running center $\mu\in\mathbb R^{512}$ is an exponential moving average with momentum 0.9, updated without gradients from both wordings of preserve pairs. The predictor is Linear--LayerNorm--ReLU--Linear of width 512, and Eq.~\ref{eq:align} uses centered opposite-wording vectors as stopped targets~\cite{simsiam,dino}.

The auxiliary loss updates the pooling head, predictor and trainable CLIP/text-conditioning path through the online features. It has no direct gradient to denoiser-only parameters through the detached weights. Registers occur after EOS and are excluded from this content pool; denoising and trajectory losses train the denoiser and registers. Recomputing attribution on the next step does not introduce a second-order path in the current step. The raw pooled cosine $\cos(u,u')$ is a diagnostic; training optimizes the centered predictor-to-target similarity. Section~\ref{app:completed_representations} evaluates the saved-head objective.

\textbf{Data-state targets.} Supporting preserve losses are
\begin{equation*}
\begin{aligned}
\mathcal L_{\mathrm{anc}}&=\langle\|f_\theta(x_t,t,c')-f_{\mathrm{ref}}(x_t,t,c)\|_M^2\rangle_{\pre},\\
\mathcal L_{\mathrm{self}}&=\langle\|f_\theta(x_t,t,c')-\mathrm{sg}(f_\theta(x_t,t,c))\|_M^2\rangle_{\pre}.
\end{aligned}
\end{equation*}
These compare predictions on noised real data. Equation~4 applies the relation target to student-visited states.

\textbf{Token roles and diagnostics.} Registers in vision transformers~\cite{darcet2024registers}, attention sinks in language models~\cite{xiao2024attentionsinks}, and padding-token analyses in text-to-image models~\cite{toker-etal-2025-padding} show that non-content positions can have computational roles. We separate the content-token pool from learned padding registers. Visual similarity alone does not establish explanatory faithfulness~\cite{saliencysanity}; the saved-head and intervention diagnostics in Section~\ref{app:completed_representations} test these links to the trained objective.

\textbf{Text interface.} For each padding position, we initialize its register with the mean over captions padded there, using the first 4,000 sorted unique original training captions. The pre-conditioning CLIP feature map, denoted $H^{\mathrm{CLIP}}$ to distinguish it from $H$, has 77 positions. We add Gaussian noise to this feature map, isotropic in standardised feature coordinates. The feature-noise loss $\mathcal L_{\mathrm{noise}}$ is the masked mean squared error between the perturbed student prediction and the unperturbed reference prediction under the original command. Its coordinatewise standard deviation $\sigma_{\mathrm{feat}}$ is half the corresponding empirical feature standard deviation. The adversarial loss $\mathcal L_{\mathrm{adv}}$ combines mean masked prediction errors to the frozen reference and stopped clean student with attention stability, evaluated under a feature perturbation obtained by three projected sign-gradient steps. The perturbation is bounded in the maximum absolute coordinate norm, scaled by empirical feature standard deviations. Top-$k_a$ attention stability retains the $k_a$ highest-attention token entries, using the SATO perturbation and attention definitions~\cite{sato}. These regularisers are shared across the matched complete recipes.

\textbf{Complete base objective.} With nonnegative coefficients $\lambda_{\mathrm{al}},\lambda_{\mathrm{anc}},\lambda_{\mathrm{s}},\lambda_n$ weighting alignment, data-state reference anchoring, data-state self-anchoring and feature noise:
\begin{equation*}
\begin{split}
\mathcal L_{\mathrm{base}}={}&\mathcal L_{\mathrm{den}}+\lambda_{\mathrm{al}}\mathcal L_{\mathrm{align}}
+\lambda_{\mathrm{anc}}\mathcal L_{\mathrm{anc}}\\
&+\lambda_{\mathrm{s}}\mathcal L_{\mathrm{self}}
+\lambda_n\mathcal L_{\mathrm{noise}}+\mathcal L_{\mathrm{adv}}.
\end{split}
\end{equation*}
Equation~5 adds trajectory anchoring to this base. The recipes also log edit-token and unchanged-token differences between detached attribution weights. Those scalar differences contribute no gradient to the parameter update. All parameters on ECHO's inference path are trained, with one fifth of the base learning rate for the CLIP tower. The benchmark configuration uses $\lambda_o=0$; matched target comparisons use the original anchor. Learned registers remain at inference, whereas the alignment heads and feature perturbations are used only during training.
\subsection{Trajectory-target derivations}
\label{app:theory}
The manuscript uses $x_t$ for a noised real-data state and $s_k$ for the motion state after $k$ sampler steps. Diffusion time $t$ and physical frame index $\ell$ are distinct. The paired command always carries a prime; subscripts $o,p$ distinguish original and paired predictions at the same state. Physical speed is denoted $\nu$, while $v_\theta$ denotes a velocity field in the generator's normalized motion-coordinate space.

\textbf{Diffusion and flow in clean-motion coordinates.} On the differentiable path of Eq.~2, solving the samplewise velocity relation together with Eq.~2 for $x_0$ gives the affine clean-motion conversion
\begin{equation*}
\begin{aligned}
\dot x_t&=\dot\alpha_t x_0+\dot\sigma_t\epsilon,\\
f_\theta(x,t,c)&=
\frac{\dot\sigma_t x-\sigma_t v_\theta(x,t,c)}
{\alpha_t\dot\sigma_t-\dot\alpha_t\sigma_t},
\end{aligned}
\end{equation*}
whenever the denominator is nonzero, where $x$ is a model input state at time $t$. For $\alpha_t=1-t$ and $\sigma_t=t$, the denominator equals one, recovering Eq.~3. Matching conditional mean velocities under squared loss gives the corresponding conditional mean clean-motion prediction because this conversion is affine. A diffusion noise predictor uses the separate noise-to-clean conversion in Sec.~3; velocity and noise predictions are distinct parameterisations, and their time-dependent conversion factors rescale prediction differences so unweighted native losses need not coincide in clean-motion coordinates.

\textbf{Proof of Proposition~3.1.} Define the anchor errors
\begin{equation*}
e_p=\|F_p-T_p\|_M,\qquad e_o=\|F_o-T_o\|_M.
\end{equation*}
Applying the reverse triangle inequality to the two response vectors and then weighted Cauchy--Schwarz to the anchor errors gives
\begin{equation*}
\begin{aligned}
\big|\,\|F_p-F_o\|_M-\delta_T\,\big|&\le e_p+e_o,\\
e_p+e_o&\le\sqrt{(\lambda_e^{-1}+\lambda_o^{-1})
(\lambda_e e_p^2+\lambda_o e_o^2)}.
\end{aligned}
\end{equation*}
Together these inequalities establish Eq.~6. For a text-independent common prediction $y$ at this state, completing the square gives
\begin{equation*}
\begin{split}
\inf_y\big\{\lambda_e\|y-T_p\|_M^2+\lambda_o\|y-T_o\|_M^2\big\}
=\frac{\lambda_e\lambda_o}{\lambda_e+\lambda_o}\delta_T^2.
\end{split}
\end{equation*}
Thus a nonzero same-state reference response prevents simultaneous elimination of both anchor errors by ignoring text. The bound governs that fixed state; correctness of the reference's response and final sampled action are separate empirical questions.

\textbf{Proof of Proposition~3.2.} Let $\mathcal N_M(F)$ extract the valid coordinates of a motion prediction $F$, flatten them and divide by $\sqrt Z$, so that its Euclidean norm matches the masked norm. Denote the frozen parameter Jacobians at $\theta_*$ by
\begin{equation*}
\begin{aligned}
\|\mathcal N_M(F)\|_2&=\|F\|_M,\\
J_p=\frac{\partial\mathcal N_M(F_p)}{\partial\theta},&\qquad
J_o=\frac{\partial\mathcal N_M(F_o)}{\partial\theta}.
\end{aligned}
\end{equation*}
Write $h=\theta-\theta_*$ for a parameter displacement. Differentiating the student branches while stopping the preserve target gives the linearised semi-gradient $Ah$ of Eq.~7; zero residual eliminates terms involving second derivatives of the Jacobians. Setting $a=J_ph$ and $b=J_oh$ with Euclidean inner product $\langle a,b\rangle=a^\top b$ gives
\begin{equation*}
\begin{aligned}
h^\top Ah
&=2w_p\|a\|_2^2-2w_p\langle a,b\rangle+2w_o\|b\|_2^2\\
&=2w_p\|a-\tfrac12 b\|_2^2+(2w_o-\tfrac12 w_p)\|b\|_2^2.
\end{aligned}
\end{equation*}
If $w_p>0$ and $w_o>w_p/4$, this quadratic form is strictly positive for any nonzero $h\in V$. Assume $V$ is nontrivial, let $A_V$ be the restriction of $A$ to $V$, and define its symmetric-part spectral margin and operator norm as
\begin{equation*}
\mu=\lambda_{\min}\!\left(\frac{A_V+A_V^\top}{2}\right)>0,\qquad
\beta_A=\|A_V\|_2.
\end{equation*}
For an ordinary gradient step of size $\eta$,
\begin{equation*}
\|h-\eta Ah\|_2^2\le
(1-2\eta\mu+\eta^2\beta_A^2)\|h\|_2^2,
\quad 0<\eta<\frac{2\mu}{\beta_A^2}.
\end{equation*}
The subspace $V$ is invariant under this linear map, and common-nullspace directions remain unchanged. The threshold is tight over arbitrary Jacobians: scalar $J_p=1/2$ and $J_o=1$ give $A=2w_o-w_p/2$, which turns negative below the threshold and vanishes at equality.

\textbf{Relation to minibatch weights.} For a per-rank microbatch of $B$ examples containing $B_p>0$ preserve pairs, each preserve example contributes $w_p=\lambda_p/B_p$ and $w_o=\lambda_o/B$ to Eq.~5. The local sufficient condition becomes
\begin{equation*}
\lambda_o>\frac{B}{4B_p}\lambda_p.
\end{equation*}
Under a single-pair convention ($B=B_p=1$, as in Full microbatches), this reduces to $\lambda_o>\lambda_p/4$, which $(\lambda_p,\lambda_o)=(4,2)$ meets. Accumulation and rank averaging apply common positive scales. Under the prescribed $80\%$ preserve mixture in the controlled Lite comparisons, the expected threshold is $1.25$, which $(\lambda_p,\lambda_o)=(4,2)$ strictly satisfies at fixed state and first-order Jacobians.

\subsection{Hyper-parameters and compute}
\label{app:hparams}
For the benchmark configuration, loss weights are $\lambda_f{=}0.5$, $\lambda_u{=}0.1$, $\lambda_{\mathrm{al}}{=}1$, $\lambda_{\mathrm{anc}}{=}4$, $\lambda_{\mathrm{s}}{=}1$, $\lambda_p{=}\lambda_e{=}4$, $\lambda_n{=}1$. The pooling head and temperature follow Eq.~\ref{eq:pool}.

For trajectory supervision we run the deployed DPM-Solver with $K{=}10$ steps and guidance 2.5. The paired instructions share a stop step drawn uniformly from $\{1,\dots,9\}$ within each worker's minibatch. Each padding position has a learned register initialized from that position's mean padding condition. The isotropic noise scale is 0.5 per-feature standard deviations. The adversarial perturbation takes three sign-gradient steps of 0.02 inside an $L_\infty$ ball of 0.05 standard deviations. Its three loss weights are 0.5 for the reference prediction, 0.25 for the clean prediction and 1 for top-$k_a$ attention stability with $k_a{=}5$.

The Table~1 benchmark uses one GPU and 32 motions per length bucket in each update. Captions with both relation types draw \edit{} with probability 0.5. The matched Lite runs in Table~2(b) use four workers with eight examples each and the five-group mixture in Section~\ref{app:joint}. Both use AdamW with learning rate $5{\times}10^{-5}$ ($10^{-5}$ for CLIP), $(\beta_1,\beta_2)=(0.5,0.9)$, weight decay $10^{-6}$, gradient clipping at 1, and EMA 0.9995. Learning rates fall tenfold at 30k of 50k updates. The single-GPU benchmark timing on an RTX~4090 was about 0.9\,s per update (12.5\,h total), with peak memory of 18.4\,GB or 8.9\,GB with activation checkpointing. Generation took 227\,ms per command.

\section{Benchmark design and evaluation}\label{supp:protocol}
To make the main-paper comparisons reproducible, this section documents the benchmark pair sources, human-caption screening and baseline training setups (Sections~\ref{app:legacydata}--\ref{app:legacytraining}). It then specifies the released and paraphrase-adapted evaluators, the 200-family video-language check and the three-seed generation protocol behind Table~1 (Sections~\ref{app:eval}--\ref{app:seedvar}).

\subsection{Benchmark data and perturbation sources}
\label{app:legacydata}
Training pairs come from the 10{,}792 training motions of ECHO: 161k LLM rewordings (lexical P1, syntactic P2, narrative P3), 177k human alternative-caption pairs, 92k rule rewordings (function-word deletion, adjacent-word swap, one typo) and 4{,}410 left/right edits. We used Google Gemini (gemini-3.1-flash-lite) and the same P1/P2/P3 definitions for training and test rewordings. The benchmark manifest lists the original caption and all three LLM rewrites for each of the 2{,}084 held-out test motions. The test set also includes human alternatives for 2{,}074 motions, the three rule rewordings, and 840 left/right edit pairs. As a \emph{zero-shot} family we use the rephrase and typo perturbations of $\delta$HumanML3D-X released with FRoM-W1~\cite{fromw1}, line-aligned to our test captions and held out from fine-tuning.

\subsection{Curation of human caption pairs}
\label{app:data}
To screen the 430k candidate preserve pairs, we measure the reference generator's one-step prediction discrepancy at the first sampler time $t_0$:
\begin{equation*}
\Delta_{\mathrm{ref}}(c,c')=1-\cos\!\big(f_{\mathrm{ref}}(s_0,t_0,c),\,f_{\mathrm{ref}}(s_0,t_0,c')\big).
\end{equation*}
The scoring script averages this discrepancy over two initial Gaussian draws, sharing $s_0$ across both captions within each draw. Among the human pairs with the largest discrepancies, one caption often describes an action that the other omits (e.g., ``raising arms up and down.'' vs.\ ``a person lifts both arms in front of his chest then swings them back down and then lifts his left arm up to his face and drops it back down''), whereas high-discrepancy LLM pairs reword the same action (``a figure makes a slide upward to the top right.'' vs.\ ``a character executes a shift upward to the top right.'').

Dropping the top 10\% highest-discrepancy human pairs improved both caption retrieval and edit flip rate in a 10k-step pilot screen by removing mismatched human annotations that describe distinct sub-actions. The Table~1 \method{} model and both joint models apply this filter, whereas the minimal-core comparison omits it.

\subsection{Benchmark baselines and validation rule}
\label{app:legacytraining}
Shared-backbone comparisons pair adapted training baselines and inference-time controls on the published ECHO Lite weights with two native humanoid systems. Below we summarize each baseline family and the validation rule that selects one checkpoint per training run.

\textbf{Baselines.} \emph{Adapted training baselines}, trained from the published ECHO weights with our pairs, 50k steps, batch 32, the same optimiser and EMA: LwF~\cite{lwf} (teacher-regularised fine-tuning), LEAF~\cite{leaf} (adversarial CLIP fine-tuning), MoCLIP~\cite{moclip} (motion-aware CLIP adaptation with tethering) and SATO~\cite{sato} (feature-space adversarial perturbation with attention stability, consistency and teacher losses, \mbox{CLIP unfrozen}). MultiAct~\cite{multiact} (attention amplification) and WINRO~\cite{winro} (initial-noise retrieval) use ECHO at inference. \emph{Native language-to-humanoid systems}: TextOp~\cite{textop} (autoregressive generation to the test length) and FRoM-W1~\cite{fromw1} (official H-GPT++ chain-of-thought LoRA on Llama-3.1-8B; retargeted to the G1 by direction-matching IK).

\textbf{Validation rule.} Checkpoint selection compares the five EMA checkpoints saved at 10k-update intervals. For each run we retain candidates with FID at most $1.10$ times its minimum validation FID, then maximize the sum of original and rewritten R@3 and the three source-specific RepCos values. Validation covers 1,630 LLM-reworded captions (two seeds), 1,628 human captions and 1,517 rule rewrites. The trainer separately logs fixed-noise reconstruction errors every 2k updates. Selection uses generated-motion metrics.

\subsection{Evaluators}
\label{app:eval}

Both Lite protocols use the released MoCLIP motion encoder for FID, FID$_D$ and RepCos, while Edit Flip measures directional shifts in limb activity. Retrieval guides validation and is reported in the detailed tables. The benchmark paraphrase-adapted tower fine-tunes on training clean/rewording pairs with InfoNCE at learning rate $10^{-5}$ for 4{,}000 updates under a clean-R@1 constraint, while the human-adapted tower excludes LLM rewordings during adaptation. Because these text towers share the same frozen motion encoder, FID and RepCos remain identical across evaluators. As Table~\ref{tab:eval} shows, the frozen released text tower loses substantial retrieval accuracy on rewordings even for real ground-truth and reused original motions, whereas paraphrase adaptation restores retrieval sensitivity to reworded clips.

\begin{appdisplay}
\captionof{table}{\textbf{Retrieval calibration on ECHO Lite.} GT scores real motions; the reuse control pairs every rewrite with ECHO's original motion.}\label{tab:eval}\tabsetup\resizebox{\linewidth}{!}{\setlength{\tabcolsep}{3.5pt}\begin{tabular}{@{}lrrrrrrrrrr@{}}
\toprule
\strideband{2}{strideTableHead}
\bfseries Evaluator   &  \multicolumn{2}{c}{\bfseries GT, original caption}   &  \multicolumn{3}{c}{\bfseries GT, rewrite captions R@3}   &  \multicolumn{3}{c}{\bfseries R@3$_P$, original ECHO motion reused}   &  \multicolumn{2}{c}{\bfseries Mean R@3$_P$} \\
\cmidrule(lr){2-3}\cmidrule(lr){4-6}\cmidrule(lr){7-9}\cmidrule(l){10-11}
\bfseries   &  \bfseries R@1   &  \bfseries R@3   &  \bfseries P1   &  \bfseries P2   &  \bfseries P3   &  \bfseries P1   &  \bfseries P2   &  \bfseries P3   &  \bfseries ECHO   &  \bfseries STRIDE \\
\midrule
\strideband{1}{strideTableStripe}
released (frozen CLIP text tower) & 0.282 & 0.475 & 0.337 & 0.397 & 0.281 & 0.311 & 0.376 & 0.276 & 0.301 & 0.321 \\
\strideband{1}{strideTableOurs}
paraphrase-adapted (ours) & \textbf{0.544} & \textbf{0.783} & \textbf{0.739} & \textbf{0.769} & \textbf{0.706} & \textbf{0.727} & \textbf{0.746} & \textbf{0.683} & \textbf{0.628} & \textbf{0.764} \\
\strideband{1}{strideTableStripe}
LLM-blind (human captions only) & \underline{0.489} & \underline{0.724} & \underline{0.570} & \underline{0.675} & \underline{0.492} & \underline{0.567} & \underline{0.658} & \underline{0.476} & \underline{0.532} & \underline{0.600} \\
\bottomrule
\end{tabular}}

\end{appdisplay}

\textbf{Adapted retrieval in the matched comparisons.} The joint comparison also evaluates a shared CLIP text tower fine-tuned on training human paraphrases paired with real motions, keeping the motion encoder fixed. Adaptation uses 373 motion clips and a separate validation split isolated from generator selection and test by motion family and normalized text, selecting update 1,800 of 4,000 under a clean-R@1 constraint. This raises validation rewording retrieval before freezing the evaluator for every generator; released-tower retrieval supplies a calibration baseline, while text-side adaptation leaves FID, FID$_D$ and RepCos unchanged.

\textbf{Overlap between splits.} The benchmark training set shares no motion families with the LLM test set, though a small subset of generic captions recurs across distinct motions. In the Full-backbone protocol, train, final selection and test splits use disjoint motion families; final-selection texts are disjoint from test, and evaluator adaptation and checkpoint selection remain isolated by both motion family and normalized caption text.

\subsection{Video-language check on synonym rewrites}\label{app:vlmcheck}
We used one P1 synonym rewrite from each of 200 nonmirrored test motion families, combining previously audited families and diagnostic cases with quotas across action type and clip duration. All eight methods generated motion from the same rewrites with seed 42. We rendered each generated G1 trajectory and its dataset reference at 50\,Hz using one fixed camera per family before tracking, widening the view on 21 families to keep all methods in frame. For each pair, WINRO selected noise separately for each caption from the same eight candidates.

Qwen3.5-9B saw the reference video first, one anonymous generated video second, and the original action caption, using 64 uniformly spaced frames with source timestamps and greedy decoding. The judge described visible motion in both clips and checked the action together with any specified direction, body side, count or order. Only a full match on visible requirements counts toward the numerator in Table~1, while all 200 families enter the denominator including uncertain verdicts.

The paired columns in Table~\ref{supp:judges} reflect binary full-match decisions. Calibration checks on the original camera views show that the judge rejects every pose-frozen failure control while accepting most identical-video controls, indicating a strict full-match threshold even for reference clips.

\subsection{Variation across generation seeds}
\label{app:seedvar}
Table~1 averages FID, FID$_D$, RepCos and Edit Flip over generation seeds 45, 46 and 47 across the same 2,084 motions and 840 edit pairs, while the VLM check and G1 tracking evaluate shared seed-42 generations. Table~\ref{supp:seedrows} disaggregates each seed: \method{} exhibits tighter cross-seed RepCos stability than ECHO and holds its ranking advantage on every individual seed. Within each seed, FID uses original-command generations and FID$_D$ and RepCos average across the three rewrite types before averaging over seeds.

\textbf{Additional measures.} For each motion, R@3 compares its paired caption with 31 distractor captions, counting a hit when at most two distractors have strictly higher cosine similarity. The sampling floor measures RepCos between two noise seeds for the same caption. Side accuracy averages individual caption-level tests of dominant limb activity, whereas paired side accuracy requires both original and edited sides to match and Edit Flip tests the direction of activity shift. Execution diagnostics cover local MPJPE, retrieval xR@3, root-path consistency xRTC~\cite{echo} and consistency on human alternatives. Table~\ref{tab:main:detailed} pairs the three-seed summary metrics with source-specific and zero-shot $\delta$HumanML3D-X breakdowns, showing that \method{} leads on LLM, human, rule and unseen perturbation families while preserving strong retrieval under both adapted and LLM-blind text towers.

\begin{appdisplay}
\captionof{table}{\textbf{Shared-backbone generation benchmark.} Rewr.$^{\mathrm{blind}}$ uses a text evaluator never trained on LLM rewrites; Floor is same-caption, different-noise RepCos. Best \textbf{bold}, second \underline{underlined}.}\label{tab:main:detailed}\tabsetup\begin{tabular*}{\linewidth}{@{\extracolsep{\fill}}lrrrrr@{}}
\toprule
\multicolumn{6}{@{}l}{\emph{(a) Quality, retrieval and edit response}} \\
\strideband{1}{strideTableHead}
\bfseries Method & \bfseries FID $\downarrow$ & \bfseries R@3 orig. & \bfseries R@3 rewrite & \bfseries R@3 blind & \bfseries Edit Flip $\uparrow$ \\
\midrule
\strideband{1}{strideTableStripe}
ECHO~\cite{echo} & 0.0374 & 0.749 & 0.628 & 0.532 & 0.902 \\
MultiAct~\cite{multiact} & 0.0389 & 0.749 & 0.629 & 0.533 & 0.902 \\
WINRO~\cite{winro} & 0.0383 & 0.767 & 0.640 & 0.546 & 0.897 \\
LwF~\cite{lwf} & \textbf{0.0308} & 0.768 & 0.649 & 0.545 & 0.892 \\
LEAF~\cite{leaf} & 0.0372 & 0.753 & 0.637 & 0.538 & \underline{0.906} \\
MoCLIP~\cite{moclip} & 0.0388 & 0.738 & 0.619 & 0.532 & \underline{0.906} \\
SATO~\cite{sato} & 0.0325 & \textbf{0.796} & \textbf{0.766} & \textbf{0.607} & 0.902 \\
\strideband{1}{strideTableOurs}
\textbf{STRIDE (ours)} & \underline{0.0322} & \underline{0.786} & \underline{0.764} & \underline{0.600} & \textbf{0.927} \\
\bottomrule
\end{tabular*}
\par\smallskip
\begin{tabular*}{\linewidth}{@{\extracolsep{\fill}}lrrrrrrr@{}}
\toprule
\multicolumn{8}{@{}l}{\emph{(b) Consistency and released $\delta$HumanML3D-X perturbations}} \\
\strideband{1}{strideTableHead}
\bfseries Method & \bfseries Floor & \bfseries LLM & \bfseries Human & \bfseries Rule & \bfseries R@3 reph. & \bfseries RepCos reph. & \bfseries RepCos typo \\
\midrule
\strideband{1}{strideTableStripe}
ECHO~\cite{echo} & 0.665 & 0.571 & \underline{0.634} & 0.827 & 0.576 & 0.467 & 0.449 \\
MultiAct~\cite{multiact} & 0.666 & 0.569 & 0.631 & 0.826 & 0.575 & 0.465 & 0.446 \\
WINRO~\cite{winro} & 0.672 & 0.511 & 0.544 & 0.755 & 0.592 & 0.415 & 0.401 \\
LwF~\cite{lwf} & 0.681 & 0.572 & 0.623 & 0.824 & 0.593 & 0.472 & 0.455 \\
LEAF~\cite{leaf} & 0.668 & 0.579 & 0.628 & 0.837 & 0.576 & 0.471 & 0.472 \\
MoCLIP~\cite{moclip} & 0.650 & 0.576 & 0.593 & 0.829 & 0.566 & 0.477 & 0.476 \\
SATO~\cite{sato} & 0.644 & \underline{0.783} & 0.624 & \underline{0.933} & \underline{0.684} & \underline{0.614} & \underline{0.636} \\
\strideband{1}{strideTableOurs}
\textbf{STRIDE (ours)} & 0.654 & \textbf{0.835} & \textbf{0.660} & \textbf{0.951} & \textbf{0.693} & \textbf{0.670} & \textbf{0.694} \\
\bottomrule
\end{tabular*}

\end{appdisplay}

\setcounter{dbltopnumber}{2}
\renewcommand{\dbltopfraction}{.80}
\section{Wording consistency and response to edits}\label{supp:targets}
The same-input on-policy distillation (OPD)~\cite{gkd} control predicts the frozen reference output for the paired command at the same sampled state. The target comparison fixes the training recipe and changes the prediction used to supervise a rewording. The component ablation asks a different question: whether a model that preserves wording changes can still switch the active limb.
\subsection{Joint multi-edit training and selected-model evaluation}
\label{app:joint}

In both the joint (Table~2(b)) and minimal-core (Table~2(c)) comparisons, we evaluate every 10k updates. The $1.10$ FID eligibility factor and five-term score in Section~\ref{app:legacytraining} select update 10,000 for all four models. Validation uses 301 LLM, 275 human and 268 rule records. OPD changes only the preserve target. The main comparison uses the released motion encoder and LLM seed 42. Table~\ref{tab:jointslices} adds seed 43 and the human and rule rewrites.

The joint pool pairs roughly 310k preserve pairs with 4k edit pairs spanning laterality, forward/backward direction, speed and gait, after excluding pairs whose motion families or caption strings overlap with final selection, test or evaluator fitting. Speed pairs swap an explicit speed adverb, while walk/run pairs preserve grammatical inflection and accepted modifiers. The prescribed training mixture samples 80\% preserve pairs and 5\% from each edit type with the shared human-pair filter retained, fixed before evaluation.

\textbf{Quality and retrieval.} Across both LLM seeds, human paraphrases and rule perturbations, \method{} raises RepCos while lowering both original-caption FID and wording distance FID$_D$ (Table~\ref{tab:jointslices}). Supervising rewordings with the student's own moving original-command prediction avoids pulling trajectories back toward the frozen reference's wording-sensitive predictions, improving consistency and visual quality together.

\begin{appdisplay}
\captionof{table}{\textbf{Joint-model results across rewrite sources.} Motion metrics use the shared released encoder.}
\label{tab:jointslices}\tabsetup\begin{tabular*}{\linewidth}{@{\extracolsep{\fill}}llrrr@{}}
\toprule
\strideband{1}{strideTableHead}
\bfseries Slice  &  \bfseries Method  &  \bfseries FID $\downarrow$  &  \bfseries FID$_D$ $\downarrow$  &  \bfseries RepCos $\uparrow$ \\
\midrule
\strideband{1}{strideTableStripe}
LLM, seed 42 & Same-input OPD & \underline{0.0349} & \underline{0.0118} & \underline{0.7825} \\
\strideband{1}{strideTableOurs}
LLM, seed 42 & \textbf{STRIDE} & \textbf{0.0333} & \textbf{0.0083} & \textbf{0.8149} \\
\strideband{1}{strideTableStripe}
LLM, seed 43 & Same-input OPD & \underline{0.0359} & \underline{0.0120} & \underline{0.7841} \\
\strideband{1}{strideTableOurs}
LLM, seed 43 & \textbf{STRIDE} & \textbf{0.0338} & \textbf{0.0082} & \textbf{0.8180} \\
\strideband{1}{strideTableStripe}
Human, seed 42 & Same-input OPD & \underline{0.0358} & \underline{0.0140} & \underline{0.6491} \\
\strideband{1}{strideTableOurs}
Human, seed 42 & \textbf{STRIDE} & \textbf{0.0330} & \textbf{0.0134} & \textbf{0.6613} \\
\strideband{1}{strideTableStripe}
Rule, seed 42 & Same-input OPD & \underline{0.0349} & \underline{0.0031} & \underline{0.9363} \\
\strideband{1}{strideTableOurs}
Rule, seed 42 & \textbf{STRIDE} & \textbf{0.0333} & \textbf{0.0026} & \textbf{0.9455} \\
\bottomrule
\end{tabular*}

\end{appdisplay}

\subsection{Minimal-core target comparison}
\label{app:litecore}
To test whether the preserve trajectory target helps without auxiliary regularisers, Table~2(c) strips \method{} and same-input OPD down to denoising and trajectory anchoring. Below we detail the setup and per-slice results in Table~\ref{tab:litecoreslices}.

\textbf{Isolation and selection.} This separate registered comparison removes alignment, padding registers, feature noise, adversarial perturbations, top-$k_a$ stability, data-state consistency, human-pair filtering and mining from both methods. It retains original/preserve denoising, frozen-teacher L1 weight 0.5 and unconditional weight 0.1, plus the trajectory terms in Sec.~3. Preserve weight 4 selects either the current original-command prediction (\method{}) or the frozen same-input prediction (OPD); edit and original trajectory weights remain 4 and 2, respectively. Training starts from the published Lite weights, with the same training implementation, examples, optimiser, seed 123 and global batch 32. Each method completes 50k updates and all five validation candidates; selection follows the rule in Section~\ref{app:joint}.

Both selected models use the same sampler, shared noise, test records and evaluators as the complete-recipe comparison. All record denominators are retained. Relative to the complete recipe, removal of human-pair filtering also changes effective pair sampling. Comparisons across recipes therefore do not isolate any single auxiliary term.

\textbf{Consistency and quality.} Even with every auxiliary regulariser and filter stripped away, \method{} raises RepCos and lowers both original-caption FID and FID$_D$ across both LLM seeds, human captions and rule perturbations (Table~\ref{tab:litecoreslices}). Gains are widest on LLM rewordings where surface wording shifts most, isolating the trajectory preserve target itself as the primary driver of wording invariance.

\begin{appdisplay}
\captionof{table}{\textbf{Minimal-core motion metrics across rewrite sources.}}
\label{tab:litecoreslices}\tabsetup\begin{tabular*}{\linewidth}{@{\extracolsep{\fill}}llrrr@{}}
\toprule
\strideband{1}{strideTableHead}
\bfseries Slice   &  \bfseries Method   &  \bfseries FID $\downarrow$   &  \bfseries FID$_D$ $\downarrow$   &  \bfseries RepCos $\uparrow$ \\
\midrule
\strideband{1}{strideTableStripe}
LLM, seed 42 & Same-input OPD & \underline{0.0353} & \underline{0.0196} & \underline{0.6943} \\
\strideband{1}{strideTableOurs}
LLM, seed 42 & \textbf{STRIDE} & \textbf{0.0323} & \textbf{0.0099} & \textbf{0.7868} \\
\strideband{1}{strideTableStripe}
LLM, seed 43 & Same-input OPD & \underline{0.0363} & \underline{0.0198} & \underline{0.6974} \\
\strideband{1}{strideTableOurs}
LLM, seed 43 & \textbf{STRIDE} & \textbf{0.0322} & \textbf{0.0095} & \textbf{0.7879} \\
\strideband{1}{strideTableStripe}
Human, seed 42 & Same-input OPD & \underline{0.0358} & \underline{0.0153} & \underline{0.6238} \\
\strideband{1}{strideTableOurs}
Human, seed 42 & \textbf{STRIDE} & \textbf{0.0320} & \textbf{0.0141} & \textbf{0.6516} \\
\strideband{1}{strideTableStripe}
Rule, seed 42 & Same-input OPD & \underline{0.0353} & \underline{0.0049} & \underline{0.9049} \\
\strideband{1}{strideTableOurs}
Rule, seed 42 & \textbf{STRIDE} & \textbf{0.0323} & \textbf{0.0033} & \textbf{0.9332} \\
\bottomrule
\end{tabular*}
\end{appdisplay}
\subsection{Trajectory ablations and paired laterality}

\label{app:pairedside}
\begin{appdisplay}
\captionof{table}{\textbf{Complete ablations and validation scores.} $S_{\mathrm{val}}$ is the five-term validation selection score; all test metrics use generation seed 42.}\label{tab:ablation:detailed}\tabsetup
\resizebox{\linewidth}{!}{\setlength{\tabcolsep}{3pt}\begin{tabular}{@{}lrrrrrrrrrr@{}}
\toprule
\strideband{1}{strideTableHead}
\bfseries Variant (50k)   &  \bfseries $S_{\mathrm{val}}$   &  \bfseries FID   &  \bfseries R@3   &  \bfseries R@3$_P$   &  \multicolumn{3}{c}{\bfseries RepCos (LLM / human / rule)}   &  \bfseries $\delta$ reph.   &  \bfseries side   &  \bfseries flip \\
\midrule
\multicolumn{11}{l}{\emph{evidence and interface terms, removed from the recipe without trajectory terms}} \\
\strideband{1}{strideTableStripe}
base: evidence + interface terms & \textbf{3.994} & 0.0338 & \textbf{0.801} & \underline{0.767} & \textbf{0.823} & \textbf{0.646} & \underline{0.947} & 0.657 & 0.865 & 0.906 \\
\quad w/o evidence alignment (pre.) & 3.953 & 0.0340 & 0.786 & 0.762 & 0.803 & 0.636 & 0.940 & 0.648 & \textbf{0.869} & 0.908 \\
\strideband{1}{strideTableStripe}
\quad w/o data-state anchoring (pre.) & 3.955 & 0.0340 & \underline{0.796} & \underline{0.767} & 0.801 & 0.619 & 0.940 & 0.637 & 0.865 & \underline{0.910} \\
\quad w/o register & \underline{3.982} & \underline{0.0333} & 0.793 & \textbf{0.768} & \underline{0.821} & 0.639 & \textbf{0.948} & \textbf{0.661} & 0.863 & 0.901 \\
\strideband{1}{strideTableStripe}
\quad w/o worst-case interface perturbation & \underline{3.982} & 0.0342 & 0.795 & 0.759 & 0.817 & 0.638 & 0.944 & \underline{0.660} & \underline{0.868} & \textbf{0.911} \\
\quad w/o isotropic interface perturbation & 3.981 & \textbf{0.0328} & 0.793 & 0.763 & 0.817 & \underline{0.641} & 0.942 & 0.654 & 0.861 & \textbf{0.911} \\
\midrule
\multicolumn{11}{l}{\emph{trajectory targets with a shared trainer}} \\
\strideband{1}{strideTableStripe}
\quad + self-anchoring (pre.) & \textbf{4.028} & \underline{0.0325} & 0.747 & \underline{0.720} & \textbf{0.860} & \textbf{0.709} & \textbf{0.962} & \textbf{0.725} & 0.769 & 0.881 \\
\quad + edit anchoring (edit) & 3.991 & 0.0329 & \textbf{0.794} & \textbf{0.767} & 0.819 & 0.640 & 0.946 & 0.650 & \underline{0.871} & \underline{0.917} \\
\strideband{1}{strideTableOurs}
\quad + both & \underline{4.013} & \textbf{0.0318} & \underline{0.787} & \textbf{0.767} & \underline{0.827} & \underline{0.659} & \underline{0.950} & \underline{0.666} & \textbf{0.874} & \textbf{0.920} \\
\midrule
\multicolumn{11}{l}{\emph{complete benchmark recipe (+ human-pair curation and data-state self-anchoring)}} \\
\strideband{1}{strideTableOurs}
\textbf{STRIDE} & 4.031 & 0.0327 & 0.786 & 0.764 & 0.832 & 0.660 & 0.951 & 0.670 & 0.877 & 0.918 \\
\bottomrule
\end{tabular}}

\end{appdisplay}

Table~\ref{tab:ablation:detailed} reports the complete 50k ablations and validation selection scores at generation seed 42, mirroring the benchmark rankings in Table~1. Removing any single interface or alignment term causes a modest local drop, whereas the trajectory target governs the central consistency--edit trade-off: self-anchoring alone maximizes wording consistency at the cost of collapsing left/right directionality, while pairing self-anchoring with edit anchoring recovers direction response and improves original-command FID. Self-only, edit-only and self + edit share the same trainer and base objective by toggling one trajectory weight, whereas the no-trajectory baseline uses its standalone trainer.

On the 840 seed-42 edits, Edit Flip can credit an activity shift that still leaves both movements on the same side. To test absolute limb correctness, we sum temporal standard deviations of the left and right joint angles to obtain $A_L, A_R$ and form the signed side index $d$ with $s=1$ for a left original command and $s=-1$ for a right original command:
\begin{equation*}
\begin{aligned}
d&=\frac{A_L-A_R}{A_L+A_R+10^{-8}},\\
\text{Edit Flip:}\;\;& s(d_o-d_e)>0,\\
\text{Paired side:}\;\;& s d_o>0 \;\;\text{and}\;\; s d_e<0.
\end{aligned}
\end{equation*}
Here $d_o$ and $d_e$ denote the original and edited side indices, and zero margins count as failures (Table~\ref{tab:pairedside}).

\begin{table}[t]
\centering
\caption{\textbf{Side correctness on 840 edits.} Single averages caption-level tests; paired requires both requested activity sides.}
\label{tab:pairedside}\tabsetup\begin{tabular*}{\linewidth}{@{\extracolsep{\fill}}lrr@{}}
\toprule
\strideband{1}{strideTableHead}
\bfseries Method   &  \bfseries Single side $\uparrow$   &  \bfseries Paired side $\uparrow$ \\
\midrule
\strideband{1}{strideTableStripe}
Base & 0.865 & 0.761 \\
+ self & 0.769 & 0.549 \\
\strideband{1}{strideTableStripe}
+ edit & 0.871 & 0.768 \\
\strideband{1}{strideTableOurs}
+ both & \underline{0.874} & \underline{0.769} \\
\midrule
\strideband{1}{strideTableOurs}
 \textbf{STRIDE} & \textbf{0.877} & \textbf{0.774} \\
SATO & 0.863 & 0.755 \\
\bottomrule
\end{tabular*}

\end{table}

\textbf{Retention of the reference response.} On edit pairs where published ECHO Lite already executes both requested sides correctly, self-anchoring alone loses roughly a third of those valid responses, whereas adding edit anchoring retains almost all of them (Sec.~5.2). Splitting reference-correct pairs at the median signed activity margin shows higher retention under joint anchoring in both weak-margin and strong-margin halves, alongside stronger paired accuracy on pairs where the frozen reference initially erred (Table~\ref{tab:pairedside}).

\subsection{Non-direction edits and\texorpdfstring{\protect\\}{ }action-recognizer calibration}
\label{app:semantics}
Beyond left/right laterality, Table~3 tests whether a single joint model follows speed and walk/run edits while keeping unedited attributes intact. The paragraphs below define the deterministic evaluation cohorts, physical speed and direction criteria, independent action recognizer and multi-edit supervision gains.

\textbf{Fixed semantic evaluation.} Before generation, we select test captions with one walking/running predicate and only explicit speed/direction modifiers or a straight-line phrase. Compound actions, objects, negation and temporal operators are excluded by a full-string grammar. For speed, existing speed modifiers are removed and the same walking instruction receives ``slowly'' or ``quickly''. For action identity, walking/running is swapped with grammatical inflection retained; all other accepted modifiers remain unchanged. This produces 76 speed pairs from 38 canonical motion families and 78 action pairs from 39. Original/mirror variants remain in one family. The action set contains 76 walk-to-run and two run-to-walk contexts, testing walk-to-run transfer.

For each of seeds 42, 43 and 44, the models and both commands share initial and per-step noise; all use the native ten-step sampler and CFG 2.5. All comparisons use this same evaluation, reported for the joint four-edit models in Table~3 of Sec.~5.3. All 2,772 motions in the three-model comparison are finite; errors would count as failed pairs, without retries or output-based filtering.

\textbf{Direct speed and unchanged attributes.} Let $X$ be a denormalised generated 38D motion of $L_X>1$ valid frames, where zero-based feature slice $X_{\ell,29:31}$ stores horizontal displacement in metres from frame $\ell$ to $\ell+1$. With frame rate $f_{\mathrm{fps}}=50$\,Hz, mean physical speed $\nu(X)$ and paired speed contrast $\Delta\nu$ between fast and slow outputs $X_{\mathrm{fast}}, X_{\mathrm{slow}}$ are
\begin{equation*}
\begin{aligned}
\nu(X)&=\frac{f_{\mathrm{fps}}}{L_X-1}\sum_{\ell=1}^{L_X-1}\|X_{\ell,29:31}\|_2,\\
\Delta\nu&=\nu(X_{\mathrm{fast}})-\nu(X_{\mathrm{slow}}).
\end{aligned}
\end{equation*}
Here $\nu$ is a scalar physical speed, distinct from the flow predictor $v_\theta$ in Eq.~3. A valid speed response requires $\Delta\nu>0$ (a tie, stationary or nonfinite trajectory fails) together with both outputs classified as walking; absolute slow/fast cutoffs did not pass calibration, so evaluation uses within-pair ordering. For walk/run edits, both endpoint labels must match their commands, and whenever forward/backward direction is explicit, the sum of aligned root-X displacements must match the requested sign in both outputs. This conjunction verifies both the edited attribute and checked unedited attributes on single-predicate locomotion prompts.

\textbf{Independent action measurement.} We train an eight-class action recognizer on real native G1 motions and their original captions (walk, run, jump, kick, punch, squat, sit and stand), excluding every generator-test family. Eligible canonical families have unsegmented single-action captions without compound context and split by fixed hash into 3,092 fitting and 870 calibration families with mirrors grouped together, supplying caption-derived supervision.

The recognizer consumes only 29 joint angles resampled to 128 normalized-phase frames, excluding text, root displacement, speed and duration. Three temporal convolutions (64, 128 and 128 channels, kernel size 5) with group normalization, GELU and mean/max pooling feed a 128-unit head, trained for 100 epochs with AdamW (initial learning rate $10^{-3}$, cosine decay, weight decay $10^{-4}$, class-weighted cross entropy, seed 20261103). Generated outputs never select weights or thresholds.

Calibration requires at least 20 independent families per class and a 95\% Wilson recall lower bound above 0.7 for both walking and running. Both classes clear this gate with strong held-out recall and precision, and non-locomotion actions are reliably rejected as neither walking nor running.

\textbf{Role of multi-edit supervision.} Training exclusively on rewordings and left/right edits narrows the unseen fast-versus-slow speed gap while retaining coarse walk/run accuracy. Adding explicit speed, gait and direction edit pairs to the joint training mixture (Section~\ref{app:joint}) restores physical speed separation and raises combined speed-and-walking accuracy in Table~3. Relation-conditioned anchoring thus extends beyond laterality once edit pairs are supplied.
\subsection{Principal paired comparisons}
\label{app:paired}

\begin{appdisplay}
\captionof{table}{\textbf{Uncertainty for the principal comparisons.} Accuracy changes in percentage points; RepCos in native units by motion family.}
\label{tab:keyuncertainty}\tabsetup
\begin{tabular*}{\linewidth}{@{\extracolsep{\fill}}llrr@{}}
\toprule
\strideband{1}{strideTableHead}
\bfseries Comparison & \bfseries Metric & \bfseries Change & \bfseries 95\% confidence interval \\\midrule
\strideband{1}{strideTableStripe}
Joint \method{} minus same-input OPD & RepCos & +0.0324 & 0.0274 to 0.0374 \\
Self + edit minus self-only & Paired side accuracy & +22.02 & 18.67 to 25.32 \\
\strideband{1}{strideTableStripe}
Joint \method{} minus ECHO Lite & Speed + walk & +2.19 & 0.43 to 4.39 \\
Joint \method{} minus ECHO Lite & Walk/run & +4.27 & 1.28 to 7.69 \\
\bottomrule
\end{tabular*}
\end{appdisplay}

The intervals in Table~\ref{tab:keyuncertainty} use 4,000 paired resamples of canonical families (1,042, 271, 38 and 39 families across the four comparisons), retaining record weights, grouping original and mirror variants, and keeping all three generation seeds within each semantic family (bootstrap seeds 20261030 and 20261103). All four headline gains stay strictly positive across resampled families.
\subsection{Reading the trajectory ablations}
\label{app:visualablations}
The following two examples expand Figure~5. All three variants use seed 42, ten sampler steps and CFG 2.5. Original and paired captions share initial and per-step noise. Camera, scale and timestamps are fixed within each comparison, with horizontal root position recentered. We chose nonmirrored clips of 150--350 frames.

\begingroup
\captionsetup{skip=3pt,hypcap=false}

\begin{appdisplay}
\appdisplaytext{The jog in Figure~\ref{fig:ablation_wording} is the first semantically equivalent pair in source order with captions of at most 110 characters, joint RepCos of at least 0.85 and a margin of at least 0.08 over each single target. Wording poses use saved generations.}
\includegraphics[width=0.86\linewidth]{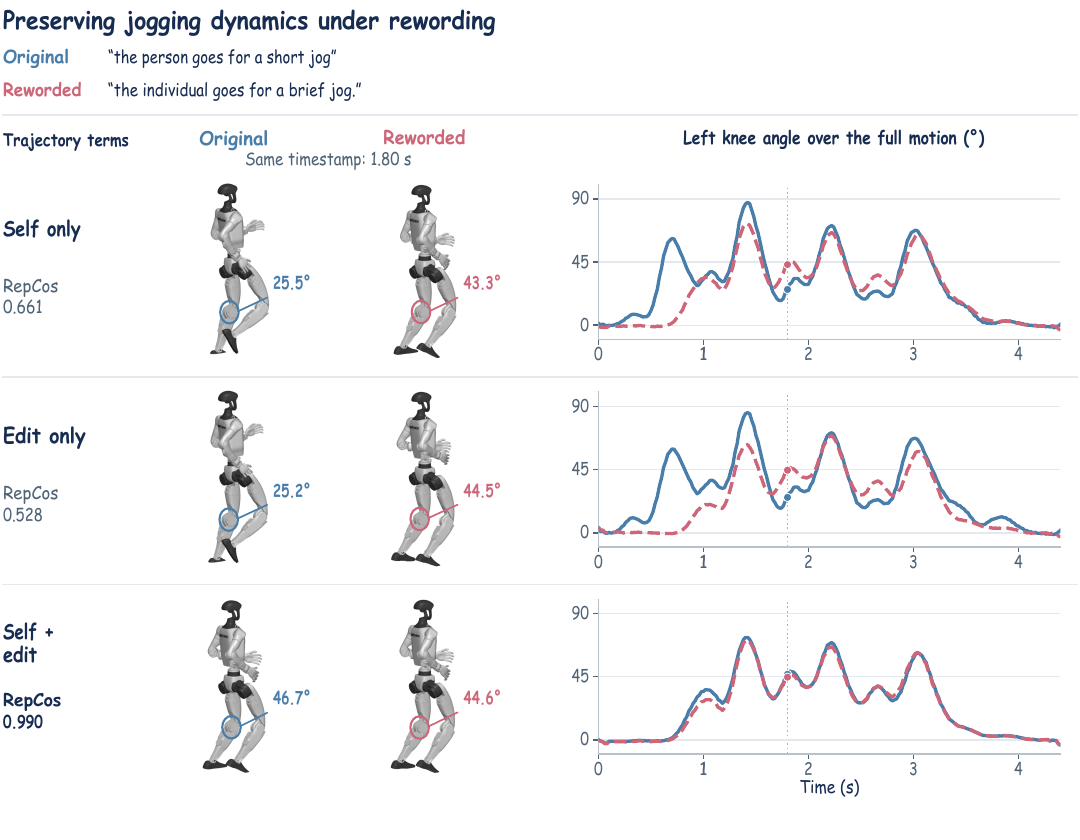}
\captionof{figure}{\textbf{Rewording a short jog (motion 006249).} Poses show the left knee at 1.80\,s. The full-motion traces compare knee flexion under the original (solid) and reworded (dashed) commands.}
\label{fig:ablation_wording}
\end{appdisplay}

\begin{appdisplay}
\appdisplaytext{For the hand edit in Figure~\ref{fig:ablation_edit}, edit-only and joint must pass both side tests and self-only must pass only the original. The original self-only margin and both joint signed margins must each be greater than 0.04. For the edit, regenerating with the saved noise and weights reproduces all six laterality indices (left/right activity difference normalized by total activity) within $2\times10^{-4}$. Table~2(a) reports results over all cases.}
\includegraphics[width=0.92\linewidth]{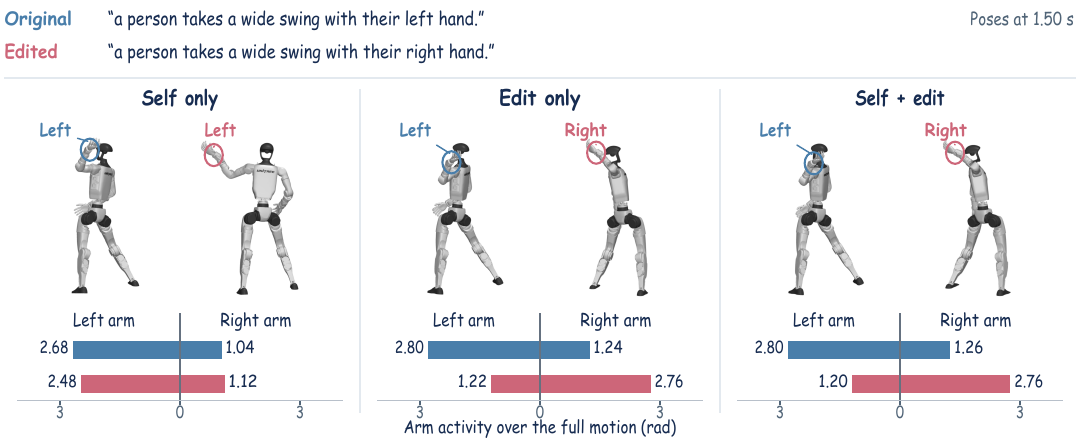}
\captionof{figure}{\textbf{Changing the swinging hand (motion 000954, pair 1957).} Poses show the generated references at 1.50\,s. The bars sum the seven joint-angle standard deviations for each arm over the full motion.}
\label{fig:ablation_edit}
\end{appdisplay}
\endgroup

\section{Generalization across corpora\texorpdfstring{\protect\\}{ }and perturbations}\label{supp:transfer}
This section documents how \method{} transfers beyond the ECHO Lite benchmark to a larger humanoid corpus and controlled text perturbations on the Full backbone. Section~\ref{app:bones} details the BONES-SEED conversion, splits and checkpoint selection behind Table~4(b), while Sections~\ref{app:completed_text}--\ref{app:pert_results} specify the eight-family perturbation protocol and per-slice Full-model metrics.

\subsection{BONES-SEED adaptation protocol}
\label{app:bones}
Table~4(b) tests target-corpus adaptation on BONES-SEED using the same \method{} training and selection procedure on an independent humanoid motion capture dataset. Below we document the 38-D corpus conversion, family-disjoint split, validation selection and rewrite-metric aggregation.

\textbf{Corpus and split.} Bones Studio's BONES-SEED dataset provides motion-capture trajectories retargeted to Unitree G1 and multiple natural-language descriptions. We convert the released G1 joint and root trajectories to ECHO's 38-D representation at 50\,fps, retaining motions of 100--490 frames. The available converted corpus contains 84,294 motions. The generator cache contains 49,874 training motions and 394 validation motions; the test set contains 720 motions, each with an original and three alternative descriptions from the source metadata. Original/mirrored motion families remain in one split. Across these disjoint motion families, 11 of the 1,724 distinct test captions also occur in \mbox{training pairs}.

\textbf{Selection.} The released evaluator scores five checkpoints on the 394 validation motions. Among checkpoints within 10\% of the best validation FID, the sum of original and rewritten R@3 and RepCos selects update 40,000. The test table scores both methods with a shared corpus-adapted evaluator at generation seed 42.

\textbf{Rewrite metrics.} We average FID$_D$ and RepCos over the three alternative descriptions. Rewritten-caption R@3 uses 32-candidate pools averaged over 20 repetitions.

For the comparison in main-paper Table~4(b), only STRIDE receives target-corpus fine-tuning. Both methods use the same held-out motions and corpus-adapted evaluator.

\subsection{Text perturbations}
\label{app:completed_text}
Table~\ref{tab:ce_text_delta} compares ECHO Full and Full \method{} under a shared protocol across all eight perturbation families. Below we specify the paired sampling controls, the seen and held-out character or framing operators, and the canonical retrieval setup.

\paragraph{Paired protocol.}
Each family uses generation seeds 42 and 43, with matched source identities, lengths, prompts, ground truth and native sampling noise. All eligible records are retained and no generation fails. Table~\ref{tab:ce_text_delta}(a) reports source records $N$ and canonical families $F$, grouping original/mirror records. Point estimates average seeds and active perturbations within each source.

\paragraph{Operators and coverage.}
These targeted invariance tests follow CheckList~\cite{checklist}. Actor keyboard substitution and character deletion modify the generic actor token, with three active variants per source. The corresponding action operators modify one action word, with one active variant. Subject deletion removes the generic subject prefix while preserving the action clause in every eligible record. Synthetic-frame deletion removes an added neutral prefix, ``In this description,'' while keeping the source command intact. This operator tests an added framing phrase; the natural neutral-deletion subset has zero eligible records. The reference caption for the framing contrast is the unframed source command.

The two additional operators preserve all characters outside one action token. Triplication changes one isolated internal character from one copy to three (e.g., \emph{walks} $\to$ \emph{waaalks}); intratoken spacing inserts one ASCII space with at least two original letters on either side (\emph{walks} $\to$ \emph{wa lks}). Their exact emitted forms and defined paired operations are absent in the eight training and validation corpora examined. The holdout applies to these exact operators and emitted forms. Broader repetition and whitespace phenomena can occur in training; actor/action typo operators serve as seen controls. All conditions evaluate held-out test motions.

\paragraph{Retrieval and interpretation.}
Canonical retrieval fixes one semantic reference per source and uses 32-item pools, 20 repeats and pool seed 42. Duplicate caption strings remain in the pools. Released and adapted text towers share the motion encoder, so their RepCos and FID measurements agree, while retrieval scales differ. Table~\ref{tab:ce_text_absolute}(b) reports both clean and perturbed retrieval levels. Eligible subsets, framing and noise streams differ across perturbation families, so their clean outputs are not pooled.

\subsection{Results by perturbation family and rewrite source}
\label{app:pert_results}
\begin{ceblock}
\captionof{table}{\textbf{Full-model results across eight text perturbations.} Panel (a) gives \method{} minus ECHO and (b) absolute means. FID uses clean commands. Retrieval ($R_c$, $R_p$) scores clean and perturbed generations against the same reference caption.}
\label{tab:ce_text_absolute}\label{tab:ce_text_delta}
{\cetab\textit{(a) Paired differences}\par\begin{tabular*}{\linewidth}{@{\extracolsep{\fill}}lcrrrr@{}}
\toprule
\strideband{1}{strideTableHead}
\bfseries Perturbation family & \bfseries $N/F$ & \bfseries $\Delta$ RepCos & \bfseries $\Delta R@3_{\rm rel}$ & \bfseries $\Delta R@3_{\rm para}$ & \bfseries $\Delta$ FID$_c$ \\
\midrule
\strideband{1}{strideTableStripe}
Action: key substitution & 1347/674 & \ceci{\textcolor{strideTableGain}{+0.2159}}{+0.1987}{+0.2334} & \ceci{\textcolor{strideTableGain}{+0.0650}}{+0.0404}{+0.0891} & \ceci{\textcolor{strideTableGain}{+0.1452}}{+0.1223}{+0.1693} & \textcolor{strideTableLoss}{+0.00407} \\
Action: character deletion & 1319/660 & \ceci{\textcolor{strideTableGain}{+0.1755}}{+0.1594}{+0.1921} & \ceci{\textcolor{strideTableGain}{+0.0263}}{+0.0001}{+0.0508} & \ceci{\textcolor{strideTableGain}{+0.1260}}{+0.1024}{+0.1499} & \textcolor{strideTableLoss}{+0.00402} \\
\strideband{1}{strideTableStripe}
Subject deletion & 1227/615 & \ceci{\textcolor{strideTableGain}{+0.1558}}{+0.1403}{+0.1716} & \ceci{\textcolor{strideTableGain}{+0.0289}}{+0.0082}{+0.0501} & \ceci{\textcolor{strideTableGain}{+0.0656}}{+0.0459}{+0.0868} & \textcolor{strideTableLoss}{+0.00390} \\
Synthetic frame deletion & 1227/615 & \ceci{\textcolor{strideTableGain}{+0.1677}}{+0.1538}{+0.1820} & \ceci{\textcolor{strideTableGain}{+0.0155}}{-0.0045}{+0.0357} & \ceci{\textcolor{strideTableGain}{+0.0448}}{+0.0254}{+0.0641} & \textcolor{strideTableGain}{-0.00116} \\
\strideband{1}{strideTableStripe}
Actor: key substitution & 1001/502 & \ceci{\textcolor{strideTableGain}{+0.2422}}{+0.2279}{+0.2565} & \ceci{\textcolor{strideTableGain}{+0.0476}}{+0.0269}{+0.0683} & \ceci{\textcolor{strideTableGain}{+0.0902}}{+0.0713}{+0.1093} & \textcolor{strideTableLoss}{+0.00494} \\
Actor: character deletion & 1001/502 & \ceci{\textcolor{strideTableGain}{+0.2049}}{+0.1899}{+0.2193} & \ceci{\textcolor{strideTableGain}{+0.0407}}{+0.0222}{+0.0599} & \ceci{\textcolor{strideTableGain}{+0.0823}}{+0.0630}{+0.1019} & \textcolor{strideTableLoss}{+0.00494} \\
\strideband{1}{strideTableStripe}
Action: character triplication & 1335/670 & \ceci{\textcolor{strideTableGain}{+0.2079}}{+0.1907}{+0.2245} & \ceci{\textcolor{strideTableGain}{+0.0837}}{+0.0577}{+0.1084} & \ceci{\textcolor{strideTableGain}{+0.1476}}{+0.1215}{+0.1738} & \textcolor{strideTableLoss}{+0.00443} \\
Action: intratoken spacing & 1333/669 & \ceci{\textcolor{strideTableGain}{+0.1991}}{+0.1820}{+0.2160} & \ceci{\textcolor{strideTableGain}{+0.0739}}{+0.0475}{+0.0999} & \ceci{\textcolor{strideTableGain}{+0.1487}}{+0.1233}{+0.1735} & \textcolor{strideTableLoss}{+0.00431} \\
\bottomrule
\end{tabular*}
\par\smallskip
\textit{(b) Absolute scores}\par\begin{tabular*}{\linewidth}{@{\extracolsep{\fill}}llrrrrrr@{}}
\toprule
\strideband{1}{strideTableHead}
\bfseries Perturbation family & \bfseries Model & \bfseries RepCos & \bfseries FID$_c$ & \bfseries $R_c^{\rm rel}$ & \bfseries $R_p^{\rm rel}$ & \bfseries $R_c^{\rm para}$ & \bfseries $R_p^{\rm para}$ \\
\midrule
\strideband{1}{strideTableStripe}
Action: key substitution & ECHO & \underline{0.5790} & \textbf{0.03940} & \textbf{0.4341} & \underline{0.3441} & \underline{0.7467} & \underline{0.6228} \\
\strideband{1}{strideTableOurs}
 & \textbf{STRIDE} & \textbf{0.7948} & \underline{0.04347} & \underline{0.4322} & \textbf{0.4090} & \textbf{0.7993} & \textbf{0.7680} \\
\strideband{1}{strideTableStripe}
Action: character deletion & ECHO & \underline{0.6232} & \textbf{0.04074} & \textbf{0.4376} & \underline{0.3752} & \underline{0.7473} & \underline{0.6446} \\
\strideband{1}{strideTableOurs}
 & \textbf{STRIDE} & \textbf{0.7987} & \underline{0.04476} & \underline{0.4230} & \textbf{0.4014} & \textbf{0.7873} & \textbf{0.7706} \\
\strideband{1}{strideTableStripe}
Subject deletion & ECHO & \underline{0.7211} & \textbf{0.04081} & \underline{0.4047} & \underline{0.3929} & \underline{0.7621} & \underline{0.7376} \\
\strideband{1}{strideTableOurs}
 & \textbf{STRIDE} & \textbf{0.8769} & \underline{0.04471} & \textbf{0.4202} & \textbf{0.4218} & \textbf{0.8070} & \textbf{0.8032} \\
\strideband{1}{strideTableStripe}
Synthetic frame deletion & ECHO & \underline{0.7421} & \underline{0.04776} & \underline{0.4080} & \underline{0.4047} & \underline{0.7421} & \underline{0.7621} \\
\strideband{1}{strideTableOurs}
 & \textbf{STRIDE} & \textbf{0.9099} & \textbf{0.04660} & \textbf{0.4217} & \textbf{0.4202} & \textbf{0.8016} & \textbf{0.8070} \\
\strideband{1}{strideTableStripe}
Actor: key substitution & ECHO & \underline{0.6687} & \textbf{0.04787} & \underline{0.4208} & \underline{0.3885} & \underline{0.7641} & \underline{0.7280} \\
\strideband{1}{strideTableOurs}
 & \textbf{STRIDE} & \textbf{0.9109} & \underline{0.05281} & \textbf{0.4390} & \textbf{0.4362} & \textbf{0.8145} & \textbf{0.8182} \\
\strideband{1}{strideTableStripe}
Actor: character deletion & ECHO & \underline{0.6943} & \textbf{0.04787} & \underline{0.4208} & \underline{0.3899} & \underline{0.7641} & \underline{0.7342} \\
\strideband{1}{strideTableOurs}
 & \textbf{STRIDE} & \textbf{0.8992} & \underline{0.05281} & \textbf{0.4390} & \textbf{0.4306} & \textbf{0.8145} & \textbf{0.8164} \\
\strideband{1}{strideTableStripe}
Action: character triplication & ECHO & \underline{0.5729} & \textbf{0.03955} & \underline{0.4332} & \underline{0.3376} & \underline{0.7495} & \underline{0.6061} \\
\strideband{1}{strideTableOurs}
 & \textbf{STRIDE} & \textbf{0.7808} & \underline{0.04398} & \textbf{0.4381} & \textbf{0.4213} & \textbf{0.8070} & \textbf{0.7537} \\
\strideband{1}{strideTableStripe}
Action: intratoken spacing & ECHO & \underline{0.5541} & \textbf{0.03969} & \underline{0.4317} & \underline{0.3358} & \underline{0.7451} & \underline{0.5880} \\
\strideband{1}{strideTableOurs}
 & \textbf{STRIDE} & \textbf{0.7532} & \underline{0.04400} & \textbf{0.4367} & \textbf{0.4098} & \textbf{0.7974} & \textbf{0.7368} \\
\bottomrule
\end{tabular*}
}
\end{ceblock}
The eight-family perturbation results in Table~\ref{tab:ce_text_absolute} average across seeds and active variants within each family. Full \method{} raises RepCos and both released and paraphrase-adapted perturbed retrieval over ECHO Full on every perturbation family, including the strictly held-out character-triplication and intratoken-spacing operators, while clean-command FID rises slightly on most slices as a small fidelity--invariance trade-off.

For the pooled Full-backbone metrics in Table~4(a), FID$_D$ compares all LLM, human and rule rewrites against their repeated originals, while R@3 uses the paraphrase-adapted text tower. Table~\ref{tab:ce_source_slices} separates each rewrite source: \method{} produces substantial consistency gains on both LLM and rule rewrites, whereas human alternative captions exhibit lower RepCos than ECHO Full because independent human captions often describe distinct actions.

\begin{ceblock}
\captionof{table}{\textbf{Full-model RepCos by rewrite source at seed 42.} LLM and rule rewrites cover 1,042 families each (human: 1,037).}
\label{tab:ce_source_slices}
{\cetab\begin{tabular*}{\linewidth}{@{\extracolsep{\fill}}lcrrr@{}}
\toprule
\strideband{1}{strideTableHead}
\bfseries Rewrite source & \bfseries $N/F$ & \bfseries ECHO RepCos & \bfseries STRIDE RepCos & \bfseries Paired $\Delta$ \\
\midrule
\strideband{1}{strideTableStripe}
LLM & 2084/1042 & \underline{0.507015} & \textbf{0.759421} & \ceci{\textcolor{strideTableGain}{+0.252406}}{+0.243108}{+0.262368} \\
Human & 2074/1037 & \textbf{0.686371} & \underline{0.608184} & \ceci{\textcolor{strideTableLoss}{-0.078187}}{-0.091911}{-0.064509} \\
\strideband{1}{strideTableStripe}
Rule & 2084/1042 & \underline{0.780105} & \textbf{0.920014} & \ceci{\textcolor{strideTableGain}{+0.139908}}{+0.131590}{+0.147853} \\
\bottomrule
\end{tabular*}
}
\end{ceblock}

\section{From generated motions to G1 execution}\label{supp:execution}
This section links generated reference trajectories to closed-loop Unitree G1 execution in simulation and on hardware. Section~\ref{app:sim} details the synchronous MuJoCo tracking setup, the per-original completion and path-consistency metrics behind Table~1 and Table~2(d), and the five physical G1 trials recorded in \texttt{real\_\allowbreak robot\_\allowbreak trials.mp4}.

\subsection{Simulation details}
\label{app:sim}
Figure~1 uses test record 227. Each row shows 12 recorded poses and 22 intermediate poses over 0--6.28\,s. Horizontal display offsets encode time while joint poses, orientations and root heights remain unchanged.

The MuJoCo~\cite{mujoco} simulator is a synchronous re-implementation of the deployment stack of the G1 tracking policy (475-D observation: future root and joint targets at offsets 0/2/4/8/16, projected gravities, angular velocity, joint-position history, previous actions; action scales 0.5 and 1.0; deployment PD gains; 100-step linear/slerp transitions; motion aligned to the current yaw). Each rollout settles for 2\,s in the default pose, transitions for 2\,s, executes the reference and holds for 0.5\,s. Failure thresholds are fixed before model evaluation. All rollouts are deterministic given the reference.

\textbf{Execution aggregation.} Table~\ref{tab:embodied:detailed} compares all shared-backbone methods and native systems on matched tracking subsets. Tracking success evaluates 400 original captions from the LLM subset; xRepCos averages the three LLM rewrite comparisons; Edit Flip covers all 840 left/right pairs before tracking, and xFlip applies the same test to the tracked 400-pair subset over the requested clip length including padded frames. For PIF, failure rates are averaged across rewrites of each successfully tracked original (three LLM or two human) and then averaged equally across eligible originals. After closed-loop tracking, \method{} leads shared-backbone methods in paraphrase-induced failure, joint tracking error, executed paraphrase consistency, root-path agreement and executed edit flip while keeping original tracking success high, showing that generation-level gains carry through whole-body control.

\begin{appdisplay}
\captionof{table}{\textbf{Robustness after shared G1 tracking.} Rankings include native systems ($^\dagger$); Table~1 ranks shared-backbone methods.}\label{tab:embodied:detailed}\tabsetup\begin{tabular*}{\linewidth}{@{\extracolsep{\fill}}lrrrrrrrrr@{}}
\toprule
\strideband{2}{strideTableHead}
\bfseries Method   &  \bfseries Succ.\,$\uparrow$   &  \bfseries PIF\,$\downarrow$   &  \bfseries MPJPE\,$\downarrow$   &  \multicolumn{2}{c}{\bfseries xR@3 $\uparrow$}   &  \multicolumn{2}{c}{\bfseries xRepCos $\uparrow$}   &  \bfseries xRTC\,$\uparrow$   &  \bfseries xFlip\,$\uparrow$ \\
\cmidrule(lr){5-6}\cmidrule(lr){7-8}
\bfseries   &  \bfseries (\%)   &  \bfseries (\%)   &  \bfseries (mm)   &  \bfseries orig   &  \bfseries rewrite   &  \bfseries LLM   &  \bfseries human   &  \bfseries   &  \bfseries \\
\midrule
\strideband{1}{strideTableStripe}
GT motions (tracked) & 96.0 & -- & 26.1 & 0.578 & -- & -- & -- & -- & -- \\
\midrule
ECHO~\cite{echo} & 90.8 & 3.8 & 32.2 & 0.597 & 0.507 & 0.611 & 0.660 & 0.574 & \underline{0.900} \\
MultiAct~\cite{multiact} & 90.8 & 3.8 & 32.2 & 0.614 & 0.507 & 0.606 & 0.655 & 0.572 & 0.887 \\
WINRO~\cite{winro} & 92.5 & 3.6 & 32.2 & \underline{0.638} & 0.521 & 0.559 & 0.593 & 0.530 & 0.873 \\
LwF~\cite{lwf} & 93.5 & 2.8 & 31.5 & 0.619 & 0.524 & 0.615 & \underline{0.665} & 0.586 & 0.877 \\
LEAF~\cite{leaf} & 91.0 & 3.6 & 32.3 & 0.600 & 0.517 & 0.612 & 0.651 & 0.578 & 0.877 \\
MoCLIP~\cite{moclip} & 90.8 & 4.6 & 32.2 & 0.611 & 0.489 & 0.605 & 0.619 & 0.575 & 0.873 \\
SATO~\cite{sato} & \underline{95.0} & 2.7 & 30.5 & \textbf{0.660} & \textbf{0.631} & \underline{0.792} & 0.644 & \underline{0.771} & 0.890 \\
\midrule
\strideband{1}{strideTableStripe}
TextOp~\cite{textop}$^\dagger$ & \textbf{97.8} & \textbf{1.5} & \textbf{20.4} & 0.251 & 0.204 & 0.679 & 0.594 & 0.435 & 0.723 \\
FRoM-W1~\cite{fromw1}$^\dagger$ & 88.0 & 7.7 & \underline{28.8} & 0.160 & 0.154 & 0.617 & 0.628 & 0.509 & 0.425 \\
\midrule
\strideband{1}{strideTableOurs}
\textbf{STRIDE (ours)} & 94.0 & \underline{2.5} & 29.8 & 0.628 & \underline{0.607} & \textbf{0.836} & \textbf{0.670} & \textbf{0.815} & \textbf{0.920} \\
\bottomrule
\end{tabular*}

\end{appdisplay}
\paragraph{G1 tracking in the controlled multi-edit experiment.}\label{app:tracking_matched}
We replayed the fixed seed-42 generations from the two models in Table~2(b) through the same G1 tracker, record IDs, captions and clip lengths as the shared-backbone benchmark: 400 LLM originals with three rewrites each and 400 human-caption originals with two alternatives each (2,800 rollouts per method, retaining all tracking failures). Because the two cohorts carry different rewrite counts per original, giving every original command equal weight defines the paired completion rate $C$ reported in Table~2(d) as
\begin{equation*}
C=\frac{1}{2}\!\left(\frac{N_{\mathrm{both,LLM}}}{1200}+\frac{N_{\mathrm{both,human}}}{800}\right)
=S_{\mathrm{orig}}(1-\mathrm{PIF}),
\end{equation*}
where $N_{\mathrm{both}}$ counts pairs in which the tracker completes both the original and rewritten motions (a failed original contributes zero) and $S_{\mathrm{orig}}$ is the original-command tracking success rate. Unlike PIF, which conditions on each model's tracked originals, $C$ evaluates all 800 original commands on a common denominator; \method{} improves paired completion over same-input OPD on both the LLM and human cohorts (Table~2(d)).

Executed root-path consistency xRTC compares the tracked horizontal root trajectory under each LLM rewrite against its original command across all 1,200 pairs including failed rollouts, scoring both arc-length-resampled path shape and total distance traveled~\cite{echo}. Evaluated on the same protocol, \method{} improves both xRTC and executed RepCos over same-input OPD, and the advantage persists whether scoring padded requested lengths or executed frames alone.

Among original commands tracked successfully by both models, discordant LLM rewrite failures favor \method{} by a three-to-one margin, while human rewrite failures are evenly split. Figure~6 visualizes the first discordant LLM example in source order from each direction, plotting the normalized tracking-error ratio
\begin{equation*}
\rho_{\mathrm{err}}(t)=\max\!\left(\frac{|z_{\mathrm{err}}(t)|}{0.25\,\mathrm{m}},\,\frac{\theta_{\mathrm{err}}(t)}{50^\circ}\right),
\end{equation*}
where failure occurs once $\rho_{\mathrm{err}}>1$. Both models track both original commands cleanly. On the hopping-and-pushing rewrite, OPD crosses the failure threshold mid-rollout and collapses toward the floor while \method{} remains upright through the full sequence. On the torso-lowering rewrite, \method{} maintains root height closely but crosses the torso-tilt limit near the end of the clip.

\paragraph{Physical G1 recordings.}
The accompanying video \texttt{real\_\allowbreak robot\_\allowbreak trials.mp4} compares ECHO Lite and \method{} side by side at $1\times$ playback across five physical Unitree G1 tasks under reworded instructions (Table~\ref{tab:robot_video}): \emph{Arm crossing}, \emph{Forward walking}, \emph{Staggering backward}, \emph{Walking with a left-hand gesture} and \emph{Bounding to the left}. \method{} crosses its forearms on Task~1, walks forward steadily on Task~2, executes backward recovery steps on Task~3, raises its left arm to shoulder height while walking on Task~4, and bounds laterally to its left on Task~5, whereas ECHO Lite spreads its arms on Task~1, stalls near the start on Tasks~2 and~3, makes a lower gesture on Task~4, and turns near its starting spot on Task~5. Figure~7 and Figure~\ref{fig:hardware-sequence} detail the first two tasks by sampling eight shared elapsed times with a fixed crop and masked background while preserving the robot's source-image position and scale. On the walking task (test motion 001120, seed 42), ECHO Lite's pre-tracking reference collapses to almost zero net horizontal travel under the reworded prompt despite normal forward travel under the original wording, whereas \method{} maintains full, consistent forward displacement across both wordings.

\begin{appdisplay}
\includegraphics[width=.89\linewidth]{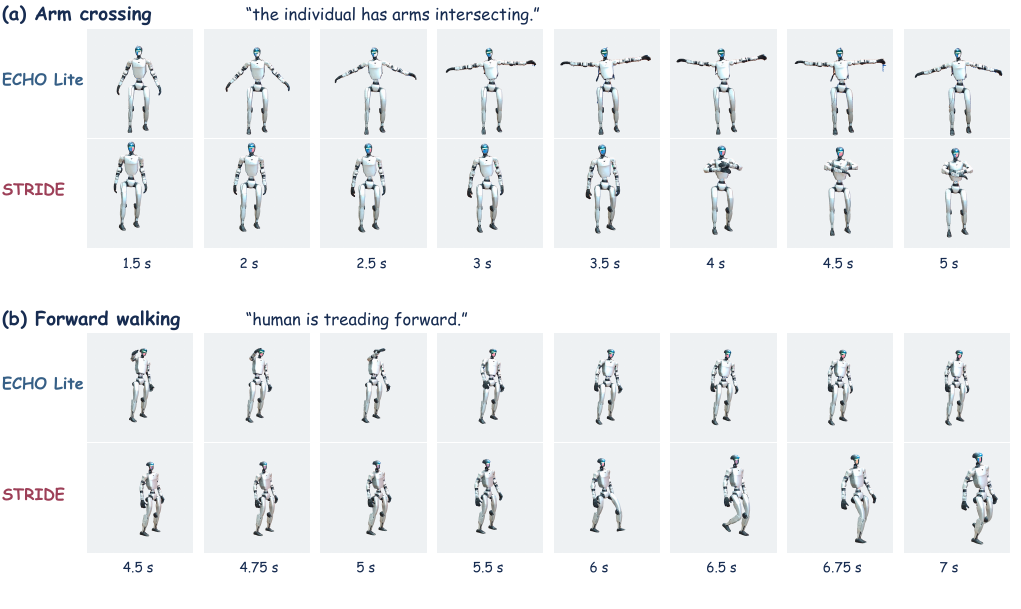}
\captionof{figure}{\textbf{Camera sequences for two reworded commands.} Each column pairs ECHO Lite and \method{} at the same elapsed video time, using a fixed camera crop that preserves foot-position changes across frames.}
\label{fig:hardware-sequence}
\end{appdisplay}

\begin{appdisplay}
\captionof{table}{\textbf{Physical Unitree G1 trials in {\mdseries\texttt{real\_robot\_trials.mp4}} (34\,s, 1080p, 30\,fps, $1\times$ playback).} Each chapter pairs ECHO Lite (left) and \method{} (right) under the same reworded instruction and controller.}\label{tab:robot_video}
\tabsetup\begin{tabular*}{\linewidth}{@{\extracolsep{\fill}}llll@{}}
\toprule\strideband{1}{strideTableHead}
\textbf{Window} & \textbf{Task chapter} & \textbf{ECHO Lite execution} & \textbf{\method{} execution}\\\midrule
\strideband{1}{strideTableStripe}
0--8\,s & Arm crossing & Spreads both arms wide to the sides & Crosses both forearms cleanly over chest\\
8--13\,s & Forward walking & Steps in place near the starting spot & Walks steadily forward with even strides\\
\strideband{1}{strideTableStripe}
13--20\,s & Staggering backward & Makes small in-place posture shifts & Sways with backward recovery steps\\
20--29\,s & Left-hand walk gesture & Walks with a low left-hand motion & Walks while raising left arm to shoulder\\
\strideband{1}{strideTableStripe}
29--34\,s & Bounding to the left & Turns near start and finishes facing away & Bounds laterally left across the floor\\
\bottomrule
\end{tabular*}
\end{appdisplay}

\setcounter{dbltopnumber}{2}
\renewcommand{\dbltopfraction}{0.85}
\renewcommand{\textfraction}{0.15}
\renewcommand{\dblfloatpagefraction}{0.95}
\Needspace{4\baselineskip}
\section{Which tokens influence the \texorpdfstring{\protect\\}{}generated motion?}\label{supp:attribution}
This section examines how motion-conditioned attribution shapes \method{}'s conditioning representations and generated trajectories. Sections~\ref{app:completed_representations}--\ref{app:attr_pool} evaluate saved-head and head-free alignment margins, compare pooling weights against local alignment-loss sensitivity, trace token attribution across timesteps, edits and full G1 rollouts, and isolate pooling rules during matched Lite training.

\subsection{Conditioning representations\texorpdfstring{\protect\\}{ }and the auxiliary objective}
\label{app:completed_representations}
\begin{ceblock}
\captionof{table}{\textbf{Saved-head representations for raw \method{}.} Margins subtract different-caption controls. Rank: entropy effective rank of centered covariance; $\sigma$: mean coordinate standard deviation.}
\label{tab:ce_raw_heads}
{\cetab\setlength{\tabcolsep}{4.8pt}\begin{tabular*}{\linewidth}{@{\extracolsep{\fill}}lrrrrr@{}}
\toprule
\strideband{1}{strideTableHead}
\bfseries Saved auxiliary representation & \bfseries Paired cosine & \bfseries Control cosine & \bfseries Margin & \bfseries Rank & \bfseries Mean $\sigma$ \\
\midrule
\strideband{1}{strideTableStripe}
Raw pool $u$ & \ceci{\textbf{0.997841}}{0.997719}{0.997956} & \ceci{0.982801}{0.982461}{0.983129} & \ceci{0.015040}{0.014696}{0.015395} & \textbf{21.83} & 0.3426 \\
Centered $u-\mu$ & \ceci{\underline{0.893515}}{0.886661}{0.899880} & \ceci{\underline{0.048049}}{0.024034}{0.072004} & \ceci{\textbf{0.845466}}{0.821114}{0.870446} & \textbf{21.83} & 0.3426 \\
\strideband{1}{strideTableStripe}
Predictor/stopped target & \ceci{0.891173}{0.884402}{0.897438} & \ceci{\textbf{0.047878}}{0.022594}{0.073562} & \ceci{\underline{0.843295}}{0.817337}{0.869692} & \underline{21.03} & 0.5836 \\
\bottomrule
\end{tabular*}
}
\end{ceblock}
The representation diagnostic evaluates all 2,084 P1 pairs (1,042 original/mirror families) in three frozen states: published ECHO EMA, \method{} EMA without auxiliary heads, and \method{} raw weights with their matching saved projector, center and predictor. Every state uses conditional evaluation mode at $t=500$, exact valid-frame scoring, and shared ground truth and noise within each pair. All records complete without failures. The saved heads and center were trained with raw weights, while EMA covers the inference model; we therefore apply the heads only to the raw state. Cosines compare captions \emph{within} one model's feature space; ECHO is never scored through \method{}'s auxiliary heads.

For each pair we report the centered predictor-to-target similarity $S$ from Eq.~\ref{eq:align}, with $v=u-\mu$ and $z=Q(v)$. The pair's alignment loss is $-S$. Table~\ref{tab:ce_raw_heads} also gives the uncentered pooled cosine. Unrelated captions sit in a narrow cone around a shared mean vector, so raw uncentered cosines remain high for both matched and unrelated pairs. Subtracting the running center $\mu$ removes that common offset and opens a wide separation between matched paraphrases and different-caption controls while preserving centered covariance, effective rank and spread $\sigma$.

All pooling rules increase paired cosine under \method{} EMA (Table~\ref{tab:ce_headfree}), and both \method{} checkpoints raise effective rank and coordinate dispersion over ECHO EMA; the centered predictor objective thus prevents dimensional collapse. Within both \method{} states, attention and uniform pooling raise unrelated-caption control cosines more strongly than GI, leaving GI with the widest paired-minus-control margin and highest effective rank. Different-caption controls verify that higher paired similarity reflects tighter grouping of equivalent wordings while preserving separation across distinct commands.

\begin{ceblock}
\captionof{table}{\textbf{Head-free comparisons in each model's 256-D conditioning space.} Pools share content support; Rank and $\sigma$ follow Table~\ref{tab:ce_raw_heads}.}
\label{tab:ce_headfree}
{\cetab\setlength{\tabcolsep}{5.5pt}\begin{tabular*}{\linewidth}{@{\extracolsep{\fill}}llrrrrr@{}}
\toprule
\strideband{1}{strideTableHead}
\bfseries Model state & \bfseries Pool & \bfseries Paired cosine & \bfseries Control cosine & \bfseries Margin & \bfseries Rank & \bfseries Mean $\sigma$ \\
\midrule
\strideband{1}{strideTableStripe}
ECHO EMA & GI & \ceci{0.6265}{0.6126}{0.6400} & \ceci{\underline{0.1583}}{0.1415}{0.1747} & \ceci{0.4682}{0.4477}{0.4892} & \textbf{30.49} & 0.5067 \\
 & Attention & \ceci{\underline{0.6486}}{0.6304}{0.6661} & \ceci{\textbf{0.0661}}{0.0401}{0.0929} & \ceci{\textbf{0.5826}}{0.5500}{0.6151} & 9.64 & 0.4226 \\
\strideband{1}{strideTableStripe}
 & Uniform & \ceci{\textbf{0.7100}}{0.6967}{0.7231} & \ceci{0.2344}{0.2143}{0.2541} & \ceci{\underline{0.4756}}{0.4544}{0.4973} & \underline{21.57} & 0.4247 \\
 \textbf{STRIDE EMA} & GI & \ceci{0.9365}{0.9328}{0.9401} & \ceci{\textbf{0.4524}}{0.4438}{0.4608} & \ceci{\textbf{0.4841}}{0.4746}{0.4937} & \textbf{49.67} & 0.5754 \\
\strideband{1}{strideTableStripe}
 & Attention & \ceci{\underline{0.9578}}{0.9547}{0.9607} & \ceci{\underline{0.5381}}{0.5290}{0.5470} & \ceci{\underline{0.4198}}{0.4104}{0.4296} & \underline{43.00} & 0.4938 \\
 & Uniform & \ceci{\textbf{0.9588}}{0.9563}{0.9612} & \ceci{0.5421}{0.5331}{0.5509} & \ceci{0.4167}{0.4074}{0.4264} & 42.32 & 0.4925 \\
\strideband{1}{strideTableStripe}
 \textbf{STRIDE raw} & GI & \ceci{0.9364}{0.9329}{0.9397} & \ceci{\textbf{0.4688}}{0.4601}{0.4773} & \ceci{\textbf{0.4676}}{0.4584}{0.4771} & \textbf{49.21} & 0.5714 \\
 & Attention & \ceci{\underline{0.9576}}{0.9545}{0.9603} & \ceci{\underline{0.5553}}{0.5462}{0.5641} & \ceci{\underline{0.4022}}{0.3928}{0.4119} & \underline{42.56} & 0.4889 \\
\strideband{1}{strideTableStripe}
 & Uniform & \ceci{\textbf{0.9585}}{0.9561}{0.9608} & \ceci{0.5593}{0.5502}{0.5681} & \ceci{0.3992}{0.3900}{0.4087} & 41.89 & 0.4876 \\
\bottomrule
\end{tabular*}
}
\end{ceblock}
\subsection{Pooling weights and local loss sensitivity}
Training weights $\alpha$ determine feature pooling. To measure local loss sensitivity, fix $\alpha$, the running center $\mu$, and the numerical values of the opposite-wording stopped targets at the observed operating point. Let $\widetilde{\mathcal L}_{\mathrm{align}}$ denote this fixed-target surrogate. For coherent scaling of one preprojection conditioning feature $H_i\mapsto(1+\delta)H_i$, the chain rule gives
\begin{align*}
\beta_i&=\left|\left.\frac{d\widetilde{\mathcal L}_{\mathrm{align}}}{d\delta}\right|_{\delta=0}\right|
=\left|\sum_d\frac{\partial\widetilde{\mathcal L}_{\mathrm{align}}}{\partial H_{id}}H_{id}\right|,\\
\widehat\beta_i&=\frac{\beta_i}{\sum_{j\in I_c}\beta_j}.
\end{align*}
When the denominator is zero, the normalized vector is defined as zero and flagged; this does not occur in the Full evaluation pairs. For the actual affine projector,
\begin{equation*}
\begin{aligned}
r&=\nabla_u\widetilde{\mathcal L}_{\mathrm{align}},\\
\nabla_{H_i}\widetilde{\mathcal L}_{\mathrm{align}}&=\alpha_iW^\top r,\qquad
\beta_i=\alpha_i|H_i^\top W^\top r|.
\end{aligned}
\end{equation*}
The projection bias is not scaled with $H_i$. Unlike the coordinatewise GI magnitude $m_i$, $\beta_i$ takes the absolute value \emph{after} the channel sum. Normalized $\beta$ measures local loss sensitivity under stopped targets, whereas recomputing the targets under a finite perturbation defines a different function. Pooling is invariant to permutation of fixed feature/weight pairs while the encoder stays order sensitive, and no cross-caption token alignment or token-position Jensen--Shannon divergence is used.

Figure~\ref{fig:ce_training_diagnostic} compares the pooling weights $\alpha$ with the local loss sensitivities $\widehat\beta$ on one paraphrase pair. Native cross-attention spreads weight broadly across articles, prepositions and pronouns alongside content terms. Both $\alpha$ and $\widehat\beta$ instead concentrate mass on the main action verbs and object phrases across the two wordings, so alignment gradients focus on the words that specify the motion while centering pulls the matched pair away from unrelated controls.

\begin{appdisplay}
\includegraphics[width=.91\linewidth]{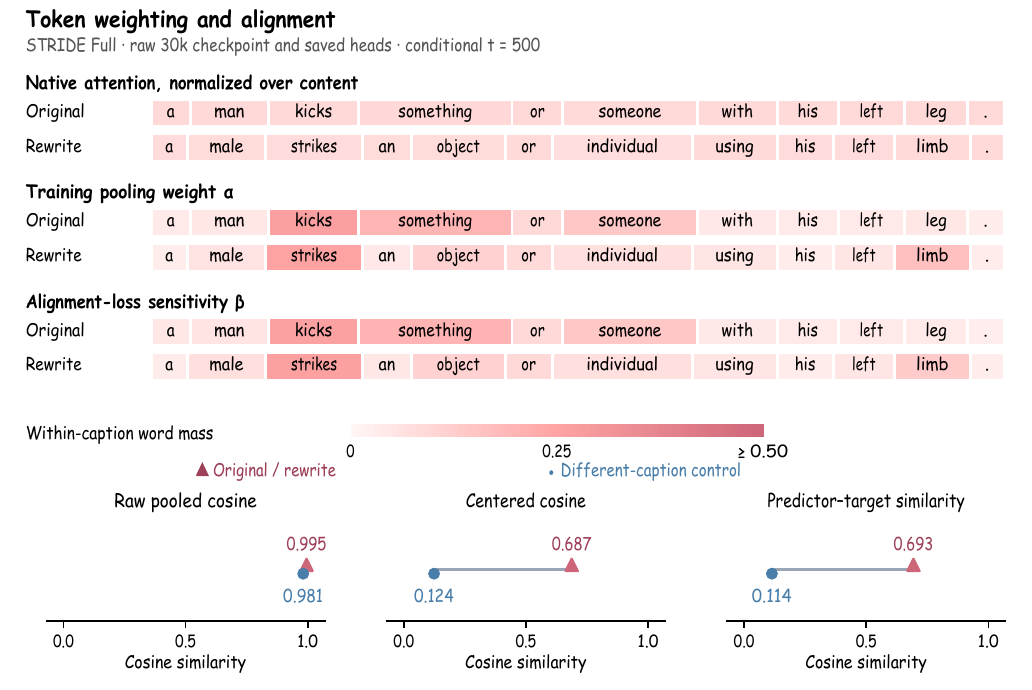}
\captionof{figure}{\textbf{Training weights and local loss sensitivity.} One $t=500$ raw \method{} pair with saved heads (population: Tables~\ref{tab:ce_raw_heads}--\ref{tab:ce_headfree}).}
\label{fig:ce_training_diagnostic}
\end{appdisplay}

\paragraph{A controlled input-embedding intervention.}
For 64 fixed nonmirror families, both deployed EMA models evaluate original and P1 captions under shared noised ground truth at $t=500$ and CFG 2.5. Each ranking scales the top 20\% of content BPE input embeddings (rounded up) by 0.8 while preserving token IDs and lengths, averaging four random masks per family across 128 finite records. This CFG input intervention differs from conditional raw-head pooling, where condition-key attention is a downstream proxy.

\begin{ceblock}
\captionof{table}{\textbf{Attenuating top 20\% content tokens by 20\% ($10^{-3}$).} GI: Grad$\times$Input; A: attention; R: random ($\Delta q<0$: worse).}
\label{tab:ce_faithfulness}
{\cetab\setlength{\tabcolsep}{4pt}\begin{tabular*}{\linewidth}{@{\extracolsep{\fill}}llrrrrrr@{}}
\toprule
\strideband{1}{strideTableHead}
\bfseries Model & \bfseries Response & \bfseries GI & \bfseries A & \bfseries R & \bfseries GI $-$ A & \bfseries GI $-$ R & \bfseries A $-$ R \\
\midrule
\strideband{1}{strideTableStripe}
ECHO & Output RMSE & \textbf{38.954} & 25.525 & \underline{26.986} & \ceci{+13.429}{+8.563}{+19.075} & \ceci{+11.967}{+8.101}{+16.625} & \ceci{-1.462}{-3.946}{+1.126} \\
ECHO & Signed $\Delta q$ & -0.573 & 0.075 & -0.404 & \ceci{-0.647}{-3.103}{+1.623} & \ceci{-0.168}{-1.704}{+1.281} & \ceci{+0.479}{-0.685}{+1.666} \\
\strideband{1}{strideTableStripe}
ECHO & $|\Delta q|$ & \textbf{5.264} & 3.153 & \underline{3.559} & \ceci{+2.111}{+0.953}{+3.319} & \ceci{+1.705}{+0.920}{+2.553} & \ceci{-0.406}{-1.112}{+0.322} \\
 \textbf{STRIDE} & Output RMSE & \textbf{8.615} & \underline{5.290} & 5.161 & \ceci{+3.325}{+1.777}{+5.567} & \ceci{+3.454}{+1.977}{+5.483} & \ceci{+0.129}{-0.749}{+1.112} \\
\strideband{1}{strideTableStripe}
 \textbf{STRIDE} & Signed $\Delta q$ & 0.056 & -0.015 & 0.062 & \ceci{+0.070}{-0.167}{+0.305} & \ceci{-0.006}{-0.194}{+0.179} & \ceci{-0.076}{-0.226}{+0.078} \\
 \textbf{STRIDE} & $|\Delta q|$ & \textbf{0.787} & 0.388 & \underline{0.462} & \ceci{+0.399}{+0.233}{+0.604} & \ceci{+0.325}{+0.168}{+0.524} & \ceci{-0.074}{-0.146}{-0.006} \\
\bottomrule
\end{tabular*}
}
\end{ceblock}

In Table~\ref{tab:ce_faithfulness}, GI-ranked attenuation produces larger output and absolute-score changes in both models, supporting its use as a local sensitivity measure. Across all three rankings, \method{} shows smaller absolute output RMSE and score shifts than ECHO because paraphrase-invariant training smooths local response to small embedding perturbations while preserving GI's lead over attention and random rankings. For \method{}, attention-ranked attenuation changes the absolute score less than random masking. The signed score measures its effect on reconstruction, whereas semantic edit probes evaluate behavioral responses to altered instructions.
\subsection{Attribution across states,\texorpdfstring{\protect\\}{ }edits and full rollouts}
\label{app:attr_sweeps}
\begin{ceblock}
\includegraphics[width=0.92\linewidth]{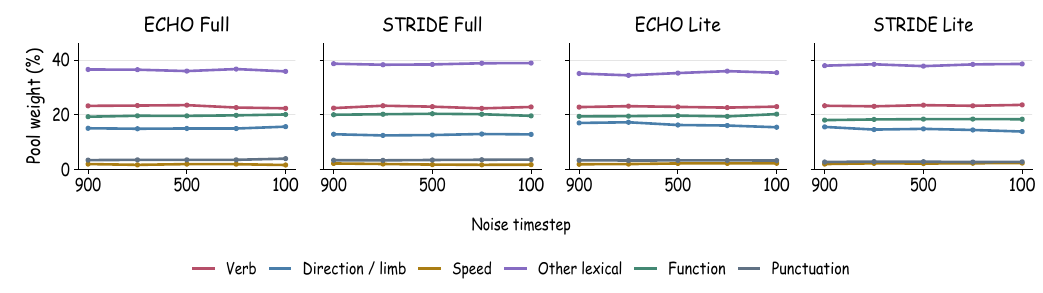}
\captionof{figure}{\textbf{Lexical words retain most pooling weight at all five noise levels.} Each caption sums to 100\% over non-special tokens. The same noise draw and motion are reused at every timestep.}
\label{fig:attr_time}
\end{ceblock}
To test whether reconstruction sensitivity tracks lexical content across diffusion noise levels and governs full generated trajectories, we sweep attribution weights across timesteps, full-corpus word roles, ten-step sampler rollouts and modifier edits. The paragraphs below define the paired evaluation cohorts and summarize the resulting motion and tracking measurements (Figures~\ref{fig:attr_time}--\ref{fig:attr_intervene}, Table~\ref{tab:attr_modifiers}).

\paragraph{Paired inputs.}
We compare ECHO and \method{} on the Full and Lite backbones. For 64 nonmirrored test families sampled without replacement (seed 20261116), the original and P1 captions share the ground-truth motion, valid length and noise draw at five timesteps. Time and modifier probes use the gradient of the conditional reconstruction score (Sec.~3.4) on these noised data states.

\paragraph{Word roles across noise levels.}
We sum BPE weights by word and average both captions within each family. A fixed lexicon identifies direction, limb and speed terms, while the NLTK perceptron tagger partitions the remaining tokens into nonauxiliary verbs, other lexical words (nouns, adjectives and adverbs), function words and punctuation. Under Eq.~\ref{eq:pool}, lexical words retain more than three quarters of total pooling weight across all five noise levels on both Full and Lite \method{}, with stable verb share from high to low noise (Figure~\ref{fig:attr_time}). Measuring per-caption mass redistribution between $t=900$ and $t=100$ via total variation,
\begin{equation*}
\mathrm{TV}=\frac{1}{2}\sum_w\big|\alpha_w(900)-\alpha_w(100)\big|,
\end{equation*}
shows that only a small fraction of weight shifts across word positions along the denoising schedule on either backbone. Because trajectory anchoring samples random intermediate steps during training, this cross-timestep stability keeps the pool focused on lexical words at every noise level without drifting to function words or punctuation.

\paragraph{Lexical and padding mass over the full test set.}
Applying the same word grouping to all 2,084 original and P1 pairs from the Full CFG diagnostic at $t=500$ with all 77 token slots in the denominator contrasts input sensitivity against downstream cross-attention. In both ECHO and \method{}, native cross-attention concentrates the majority of its mass on padding slots and assigns less than a tenth to lexical words. Reconstruction Grad$\times$Input reverses that allocation: lexical words receive roughly seven tenths of total sensitivity mass while padding receives under two percent, producing an order-of-magnitude higher lexical-to-nonlexical ratio.

\paragraph{Intervention through the sampler and G1.}
\begin{ceblock}
\includegraphics[width=0.88\linewidth]{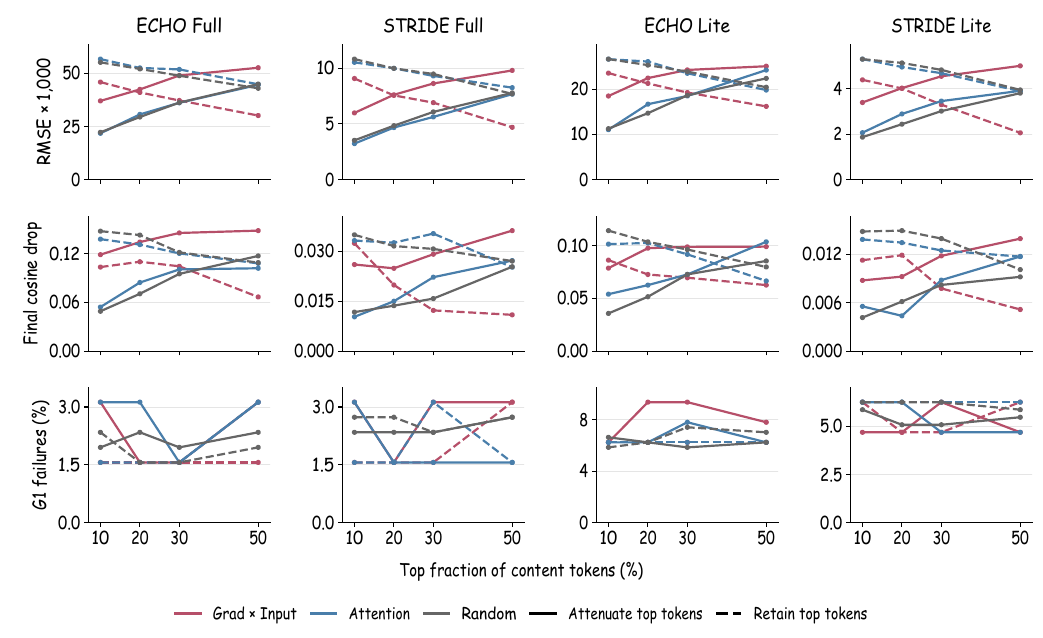}
\captionof{figure}{\textbf{Token scaling changes generated motion more consistently than tracking failure.} Solid curves attenuate the top-ranked tokens. Dashed curves retain those tokens and attenuate the rest. Each panel uses its own vertical scale.}
\label{fig:attr_intervene}
\end{ceblock}
\begin{table*}[t]
\centering
\caption{\textbf{Modifier sensitivity under matched PRE and EDIT captions.} Each side or speed cohort contains 64 original trajectories. The reconstruction target is held fixed across all three captions.}
\label{tab:attr_modifiers}
{\cetab\small\renewcommand{\arraystretch}{1.15}\begin{tabular*}{\textwidth}{@{\extracolsep{\fill}}llrrr|rrr@{}}
\toprule
\strideband{2}{strideTableHead}
 & & \multicolumn{3}{c}{Pooling mass (\%)} & \multicolumn{3}{c}{Sum of gradient norms ($\times 10^3$)} \\
\cmidrule(lr){3-5}\cmidrule(lr){6-8}Variant & Modifier & Original & PRE & EDIT & Original & PRE & EDIT \\
\midrule
\strideband{1}{strideTableStripe}
Self only & Left/right & 5.11 & 4.15 & 5.57 & 8.04 & 13.13 & 9.08 \\
 & Speed & 26.06 & 33.32 & 27.73 & 17.20 & 42.53 & 22.90 \\
\strideband{1}{strideTableStripe}
Edit only & Left/right & 5.13 & 3.52 & 5.12 & 7.48 & 11.88 & 8.42 \\
 & Speed & 27.76 & 31.69 & 29.01 & 15.59 & 43.86 & 23.44 \\
\strideband{1}{strideTableStripe}
Self + edit & Left/right & 4.95 & 3.81 & 5.42 & 7.53 & 10.92 & 8.74 \\
 & Speed & 26.09 & 31.29 & 27.38 & 16.03 & 41.85 & 23.97 \\
\bottomrule
\end{tabular*}
}
\end{table*}
For each of the 64 P1 captions we rank content tokens once at $t=500$ using the ground-truth-conditioned reconstruction score with CFG 2.5, as in Table~\ref{tab:ce_faithfulness}. For each tested budget fraction $p$, we scale either the top-ranked content embeddings or their complement by 0.8 before CLIP positional encoding, keeping token IDs and BOS/EOS fixed.

Within each caption we reuse the initial noise and all ten stochastic solver draws (seed 42), comparing each intervened rollout against the model's own clean output on one-step RMSE, motion-encoder cosine drop and shared G1 tracking failure. Four random orderings are averaged per family across 64 clean and 3,072 intervened rollouts per model, with every rollout kept in the denominator.

Attenuating the top-ranked tokens perturbs generated motion more strongly under Grad$\times$Input than under attention, though paired uncertainty at the 20\% budget remains wide. The complementary retention test separates the two rankings cleanly: keeping the top 30\% of tokens intact while attenuating the rest preserves much closer motion similarity under Grad$\times$Input than under attention (Figure~\ref{fig:attr_intervene}). Clean G1 tracking failures stay rare across all four models, and intervened failure rates depend on both model and budget without mirroring motion-cosine shifts monotonically. Whereas pose similarity degrades smoothly with embedding perturbation, closed-loop tracking failure is a threshold event governed by torso tilt and root height limits, so moderate motion drift can shift embedding similarity while staying inside the controller's stability envelope.

\paragraph{Modifier counterfactuals.}
At $t=500$, we score the original, P1 and edited captions against the same original motion and noise across 64 nonmirrored families per cohort: side edits join the benchmark left/right manifest on motion ID and exact original text, while speed edits swap a single rate adverb (\emph{slowly}, \emph{quickly} or \emph{rapidly}) for its opposite. Table~\ref{tab:attr_modifiers} evaluates the three Lite variants from Table~2(a), summing pooling mass and input-gradient norms over matching modifier words (one paraphrase lacks an explicit speed word and contributes zero). Under both self-only and joint anchoring, side-word pooling mass rises noticeably from paraphrase captions to left/right edits with comparable relative shifts across variants. Both models therefore register the edited modifier at the text input; because paired-side accuracy improves sharply only once edit trajectory anchoring is added (Table~\ref{tab:pairedside}, Sec.~5.2), input-level modifier sensitivity alone is insufficient without trajectory supervision that preserves the edited motion.

\paragraph{What the intervention identifies.}
The ranking uses the reconstruction gradient for a known reference motion and measures sensitivity to the chosen embedding perturbation at fixed noise. A gain in word mass alone cannot pinpoint the token responsible for a specific tracking failure; testing such a causal link on the physical walking example would require intervening directly on the verb under shared solver noise and measuring the resulting displacement change.

\subsection{Pooling during matched Lite training}
\label{app:attr_pool}
Three runs start from the published ECHO Lite weights under identical joint trajectory and interface objectives, differing only in the detached pooling weights passed to the preserve-pair alignment head: uniform weights over content tokens, native key-softmax cross-attention averaged over layers, heads and channels and renormalized on content positions, or reconstruction Grad$\times$Input from Eq.~\ref{eq:pool}. All three runs train identical projection and predictor architectures.

Each run uses one GPU, batch 32, seed 123 and the same 320,000 sampled training records in identical order, evaluating EMA weights after 10,000 updates. All three models generate motions for the same 2,084 originals and their P1/P2/P3 rewrites with shared solver noise across all ten steps (CFG 2.5, seed 42). RepCos and FID$_D$ average the three rewrites against original generations within each model, and R@3 retrieves each generated motion's own caption from 32-candidate pools under the released evaluator.

Across all 1,042 canonical motion families, Grad$\times$Input pooling achieves the strongest RepCos, lowest wording distribution distance FID$_D$ and highest original and rewritten retrieval among the three pooling rules (Table~\ref{tab:lite_pool_training}). Paired family intervals separate Grad$\times$Input cleanly from uniform pooling on RepCos, while the smaller gain over attention pooling still overlaps zero at this 10k-update horizon. Because uniform pooling weights function words equally with action verbs and cross-attention inherits downstream attention biases, steering the alignment head with reconstruction sensitivity focuses text-encoder updates on motion-bearing words while shared trajectory anchoring supplies the primary global invariance signal.

\end{document}